%% file: main.tex
\documentclass{article}

\PassOptionsToPackage{numbers, compress}{natbib}

\usepackage[final]{neurips_2024}

\usepackage[utf8]{inputenc} 
\usepackage[T1]{fontenc}    
\usepackage{url}            
\usepackage{booktabs}       
\usepackage{amsfonts}       
\usepackage{nicefrac}       
\usepackage{microtype}      
\usepackage{xcolor}         
\usepackage{float}

\usepackage{amsmath}
\usepackage{amssymb}
\usepackage{booktabs}
\usepackage{multicol,multirow}
\usepackage{setspace}
\usepackage{makecell}
\usepackage{tabularray}
\usepackage{pifont}
\usepackage{wrapfig}

\usepackage{color}
\usepackage{caption, subcaption, multirow, overpic, textpos}
\usepackage{bbm}
\usepackage{titletoc}
\usepackage{graphicx}

\usepackage{colortbl}
\usepackage{tcolorbox}

\usepackage{epigraph}
\newlength\savewidth
\newcommand{\tablestyle}[2]{\setlength{\tabcolsep}{#1}\renewcommand{\arraystretch}{#2}\centering\footnotesize}
\renewcommand{\paragraph}[1]{\vspace{1.25mm}\noindent\textbf{#1}}

\newcolumntype{x}[1]{>{\centering\arraybackslash}p{#1pt}}
\newcolumntype{y}[1]{>{\raggedright\arraybackslash}p{#1pt}}
\newcolumntype{z}[1]{>{\raggedleft\arraybackslash}p{#1pt}}

\definecolor{deemph}{gray}{0.6}

\definecolor{baselinecolor}{gray}{.9}
\newcommand{\baseline}[1]{\cellcolor{baselinecolor}{#1}}


\usepackage[pagebackref,breaklinks,colorlinks]{hyperref}
\usepackage[capitalize]{cleveref}
\crefname{section}{Sec.}{Secs.}
\Crefname{section}{Section}{Sections}
\Crefname{table}{Table}{Tables}
\crefname{table}{Tab.}{Tabs.}

\definecolor{Gray}{gray}{0.9}
\definecolor{LightCobaltBlue}{RGB}{143,170,220}
\definecolor{Amber}{RGB}{255,102,0}
\definecolor{BlackOlive}{RGB}{59,56,56}
\definecolor{AlloyOrange}{RGB}{197,90,17}
\definecolor{B'dazzledBlue}{RGB}{47,85,151}
\definecolor{DarkBlue}{RGB}{72, 116, 203}
\definecolor{DarkGreen}{RGB}{106, 169, 63}

\usepackage[accsupp]{axessibility}  

\title{SimGen: Simulator-conditioned Driving Scene Generation}

\author{
Yunsong Zhou$^{1, 2*}$~ 
Michael Simon$^{1}$ ~
Zhenghao Peng$^{1}$ ~
Sicheng Mo$^{1}$ \\
[1mm]
\textbf{Hongzi Zhu}$^{2}$ ~
\textbf{Minyi Guo}$^{2}$ ~
\textbf{Bolei Zhou}$^{1}$ \\
[2mm]
$^1$~University of California, Los Angeles ~
$^2$~Shanghai Jiao Tong University \\ 
[2mm]
}

\begin{document}

\maketitle
\vspace*{-2em}
\centerline{\large \url{https://metadriverse.github.io/simgen/}}

\vspace{-4mm}
\begin{figure}[h]
    \centering
    \includegraphics[width=1\linewidth]{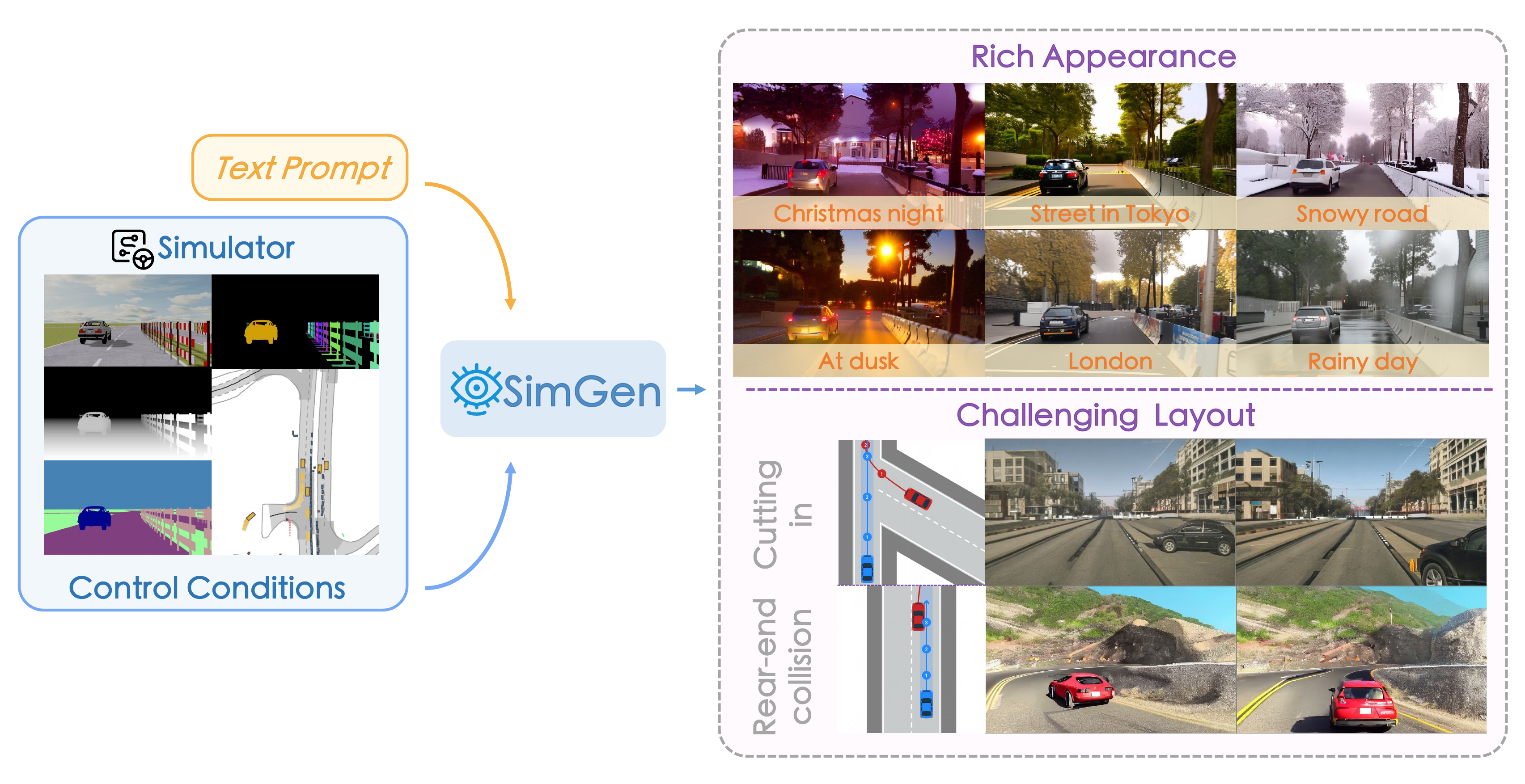}
    \caption{
    %
    \textbf{SimGen} is a \textit{controllable} scene generation paradigm conditioned on a simulator. It learns from real-world and simulated data and then can generate diverse driving scenes based on the simulator's control conditions and text prompt.
    }
    \label{fig:teaser}
\end{figure}


\begin{abstract}

Controllable synthetic data generation can substantially lower the annotation cost of training data. Prior works use diffusion models to generate driving images conditioned on the 3D object layout. However, those models are trained on small-scale datasets like nuScenes, which lack appearance and layout diversity. Moreover, overfitting often happens, where the trained models can only generate images based on the layout data from the validation set of the same dataset. In this work, we introduce a simulator-conditioned scene generation framework called SimGen that can learn to generate diverse driving scenes by mixing data from the simulator and the real world. It uses a novel cascade diffusion pipeline to address challenging sim-to-real gaps and multi-condition conflicts. A driving video dataset DIVA is collected to enhance the generative diversity of SimGen, which contains over 147.5 hours of real-world driving videos from 73 locations worldwide and simulated driving data from the MetaDrive simulator. SimGen achieves superior generation quality and diversity while preserving controllability based on the text prompt and the layout pulled from a simulator. We further demonstrate the improvements brought by SimGen for synthetic data augmentation on the BEV detection and segmentation task and showcase its capability in safety-critical data generation. 

\end{abstract}

\let\thefootnote\relax\footnotetext{\hspace*{-0.39em}\textsuperscript{*}The work was done when YZ was a visiting student at UCLA.}

\section{Introduction}

A high-quality and diverse training data corpus is crucial for autonomous driving research and development.
%
However, it is costly and laborious to annotate the data. 
%
Synthetic data generation is a promising alternative to harvest annotated training data, which brings realistic images and notable performance improvements across tasks like object detection~\cite{chen2023integrating} and semantic segmentation~\cite{wu2024datasetdm}. Besides the \textit{realism} of the generated images, there are two necessary conditions to consider for a practical synthetic data generator for autonomous driving: 
1) Appearance diversity, which ensures the synthetic data can cover a spectrum of weather, environmental, and geographical conditions. 2) Layout diversity, namely the distribution of objects, should cover different traffic scenarios, including safety-critical situations that are rare to collect in the real world. 

Recent diffusion-based generative models show promising results to generate realistic driving images from text prompts~\cite{yang2024generalized}, BEV road maps~\cite{swerdlow2024street}, and object boxes~\cite{gao2023magicdrive, wang2023drivedreamer,wen2023panacea,yang2023bevcontrol}.
Despite generating coherent images, these attempts lack the generalizability of generating new and diverse real-world appearances and traffic scenarios due to data limitations. 
%
They are confined to learning on small-scale datasets \cite{jia2023adriver,kim2021drivegan,lu2023wovogen,wang2023driving} with limited scenarios such as only urban streets~\cite{blattmann2023align} or restricted weather conditions~\cite{qi20193d}.
%
In addition, the driving behaviors in the available driving datasets like nuScenes are tedious and lack complex or safety-critical situations. 
%
%
Another option for collecting synthetic data is from driving simulators, which can effortlessly generate scenes encompassing various behaviors with its physics and graphics engines~\cite{dosovitskiy2017carla,li2022metadrive,richter2016playing,ros2016synthia,sun2022shift}. Simulators also provide accurate control over all objects and their spatial locations, thus can easily generate a huge amount of traffic layout maps. 
%
%
%
However, open-source simulators usually only contain a limited amount of 3D assets, and they lack a realistic visual appearance. Thus, the models trained on simulator-generated data can easily overfit, also known as the Simulation to Reality (Sim2Real) gap.


%
%



We take the best of two worlds by integrating the data-driven generative models with a simulator to obtain both the appearance diversity of real-world data and the layout controllability of simulated data. To this end, we introduce \textbf{SimGen}, a simulator-conditioned diffusion model, which follows the layout guidance from the simulator and rich text prompts to generate diverse driving scene images. 
One na\"ive approach is to guide an image generation model with the depth and semantic images from the simulator via training a control branch through ControlNet~\cite{zhang2023adding}.
Yet, as the simulator has limited assets and cannot fully capture the variations in the real world, the simulated conditions and the underlying real-world conditions that guide a diffusion model to generate real-world images might have conflicts.
%
To tackle this, SimGen adopts a cascade design.
The model first injects noise-added simulated conditions such as depth and semantic images into the intermediate sampling process of a pre-trained text-to-real-condition diffusion network. 
The network then converts simulated conditions into more realistic conditions via continuous denoising, free of additional training on simulated conditions beyond this diffusion network.
After that, a second diffusion module utilizes an adapter to integrate multimodal conditions and uses masks to filter conflicting data.
SimGen thus achieves outstanding generation quality and diversity while preserving layout controllability by connecting with the simulator.

We construct a dataset called \textbf{DIVA} to obtain the appearance and layout diversity of the training data.
DIVA comprises two parts: the web data and the synthesized data from the simulator.
On the one hand, web data covers a worldwide range of geography, weather, scenes, and traffic elements, preserving the appearance diversity of a wide range of traffic participants.
We design a data curation pipeline to collect and label YouTube driving videos.
On the other hand, virtual driving videos with the traffic flow replayed from trajectory datasets or generated by a safety-critical scenario generator~\cite{zhang2023cat} are collected from a driving simulator~\cite{li2022metadrive}.
In short, \textit{DIVA dataset blends real-world appearances and virtual layouts}, consisting of 147.5 hours of \textbf{D}iverse \textbf{I}n-the-wild and \textbf{V}irtual driving dat\textbf{A}.

We summarize our contributions as follows:
1) a novel controllable image generation model SimGen incorporating a driving simulator to generate realistic driving scenarios with appearance and layout diversity;
2) a new dataset DIVA containing massive web and simulated driving videos that ensures diverse scene generation and advances simulation-to-reality research;
3) SimGen improves over counterparts like BEVGen~\cite{swerdlow2024street}, MagicDrive~\cite{gao2023magicdrive}, Panacea~\cite{wen2023panacea}, DrivingDiffusion~\cite{li2023drivingdiffusion}, \textit{i.e.}, in terms of image quality, diversity, and controllability of scene generation.

\section{Appearance Diversity and Layout Diversity from DIVA Dataset}
\label{sec:diva}

We introduce a large-scale DIVA dataset containing diverse driving scenes in the real world and the simulation. It facilitates the training of generative models and tackles the simulation-to-reality (Sim2Real) challenge.
\cref{tab:DIVA} displays the statistics, composition, and annotation of the data, which comprises about 147.5 hours of driving videos.
The data is collected from a vast corpus of high-quality YouTube driving videos and simulation environments
in the MetaDrive simulator~\cite{li2022metadrive}.
%
We use DIVA-Real and DIVA-Sim to denote the web data downloaded from YouTube and the data from the MetaDrive simulator, respectively. 
%
Comparisons with other datasets, license, and privacy considerations are detailed in \cref{sec:sup_diva}.

\vspace{-3pt}
\subsection{DIVA-Real: Appearance Diversity in Web Data}
\label{section:diva-real}
\vspace{-3pt}

\smallskip
\noindent 
\textbf{Collecting web videos.}
%
%
As shown in \cref{fig:diva} (left), to streamline the process and minimize manual effort, we begin by searching for relevant keywords on YouTube to identify a batch of driving video channels.
The videos are downloaded from these identified YouTube channels. 
We filter out unsuitable videos based on their length and resolution and proceed to download the appropriate ones.
This yields hundreds of first-person driving videos, each with an average duration of one hour.
Next, we sample the videos into frames at 10Hz, excluding the initial and final 30 seconds to eliminate user channel information. 
This process yields over 4.3 million frames, awaiting further data cleaning.

\begin{wrapfigure}{r}{0.6\textwidth}
\begin{minipage}[b]{0.6\textwidth}
\vspace{-11pt}
\captionof{table}{\textbf{Comparing DIVA with relevant datasets on scale, diversity, and annotations.}
$^*$: perception subset.
$^+$: including procedural generation~\cite{li2022metadrive} and safety-critical~\cite{zhang2023cat} data.
\texttt{Cts}: countries;
\texttt{Seg}: segmentation;
\texttt{Virt}: virtual image.
%
}
\vspace{-5pt}
\begin{center}
\tablestyle{2.0pt}{1.05}
\scriptsize
\setlength{\tabcolsep}{1mm}{
\begin{tabular}{lcccccccc}
\toprule
   \multirow{2}{*}{Dataset}      & Time & \multirow{2}{*}{Frames} & \multirow{2}{*}{Cts.} & \multirow{2}{*}{Cities} & \multicolumn{4}{c}{{\scriptsize{Annotations}}} \\
   \cmidrule(r){6-9} 
   & (hours) & & & & {\scriptsize{Text}} & {\scriptsize{Depth}} & {\scriptsize{Seg.}} & {\scriptsize{Virt.}}\\
  \midrule
 KITTI~\cite{geiger2015kitti}             & 1.4              & 15k               & 1                    & 1                &    & \ding{51}& \ding{51}  &    \\
 CityScapes~\cite{cordts2016cityscapes}        & 0.5              & 25k               & 3                    & 50               &   & & \ding{51}     &     \\
 Waymo$^*$~\cite{sun2020scalability}        & 11               & 390k              & 1                    & 3                &        & &  \ding{51} &   \\
 Argoverse 2$^*$~\cite{wilson2023argoverse}       & 4.2              & 300k              & 1                    & 6                &     & &   &     \\
 nuPlan$^*$~\cite{caesar2021nuplan}            & 120              & 4.0M              & 2                    & 4                &     & &    &     \\
 Honda-HAD~\cite{kim2019grounding}         & 32               & 1.2M              & 1                    & -                &       \ding{51}   & &  &  \\ 
 nuScenes~\cite{caesar2019nuscenes}          & 5.5              & 241k              & 2                    & 2                &     &  & \ding{51}   &     \\ \midrule
 \baseline{DIVA-Real}  & \baseline{120}              & \baseline{4.3M}              & \baseline{19}                   & \baseline{71}               &   \baseline{\ding{51}} & \baseline{\ding{51}} & \baseline{\ding{51}}     &   \baseline{}  \\
 \baseline{DIVA-Sim} &  \baseline{27.5$^{+}$}              & \baseline{998k$^{+}$}              & \baseline{3}                    & \baseline{5}                &    \baseline{\ding{51}} &    \baseline{\ding{51}} &    \baseline{\ding{51}}   & \baseline{\ding{51}}    \\ 
 \baseline{\textbf{DIVA (All)}}              &    \baseline{\textbf{147.5}}              &    \baseline{\textbf{5.3M}}               &           \baseline{\textbf{22}}           &      \baseline{\textbf{76}}            &      \baseline{\ding{51}} & \baseline{\ding{51}} & \baseline{\ding{51}}  & \baseline{\ding{51}}     \\ \bottomrule
\end{tabular}
}
\end{center}
\vspace{-5pt}
\label{tab:DIVA}
\end{minipage}
\end{wrapfigure}

\smallskip
\noindent 
\textbf{Data cleaning and autolabeling.}
Data cleaning is vital for ensuring data quality, but manual inspection of each image is impractical.
Inspired by~\cite{yang2024generalized}, we implement an automated data-cleaning workflow
to expedite the process.
With the remarkable image understanding capabilities of the vision-language model (VLM), \textit{i.e.} LLaMA-Adapter V2~\cite{gao2023llamaadapterv2}, we are able to conduct the quality checks via VLM with a checklist including criteria such as non-front view, video transition, black screens, \textit{etc}, to identify nonconforming images.
%
%
Driving videos are chunked into five-frame batches.
For each batch, the VLM chooses and assesses a random image; if this single image fails to pass checks, the entire batch of five frames will be discarded.
%
%
%
In the autolabeling process, pre-trained models for various tasks,
including BLIP2-flant5~\cite{2020t5}, ZoeDepth~\cite{bhat2023zoedepth}, and Segformer~\cite{xie2021segformer}
, are used to generate annotations of text, depth, and semantic segmentation, respectively.
Eventually, over 120 hours of driving videos with rich annotations are collected.
%


\vspace{-3pt}
\subsection{DIVA-Sim: Layout Diversity from the Simulator}
\label{section:diva-sim}
\vspace{-3pt}

\begin{figure}[t]
    \centering
    \includegraphics[width=1\linewidth]{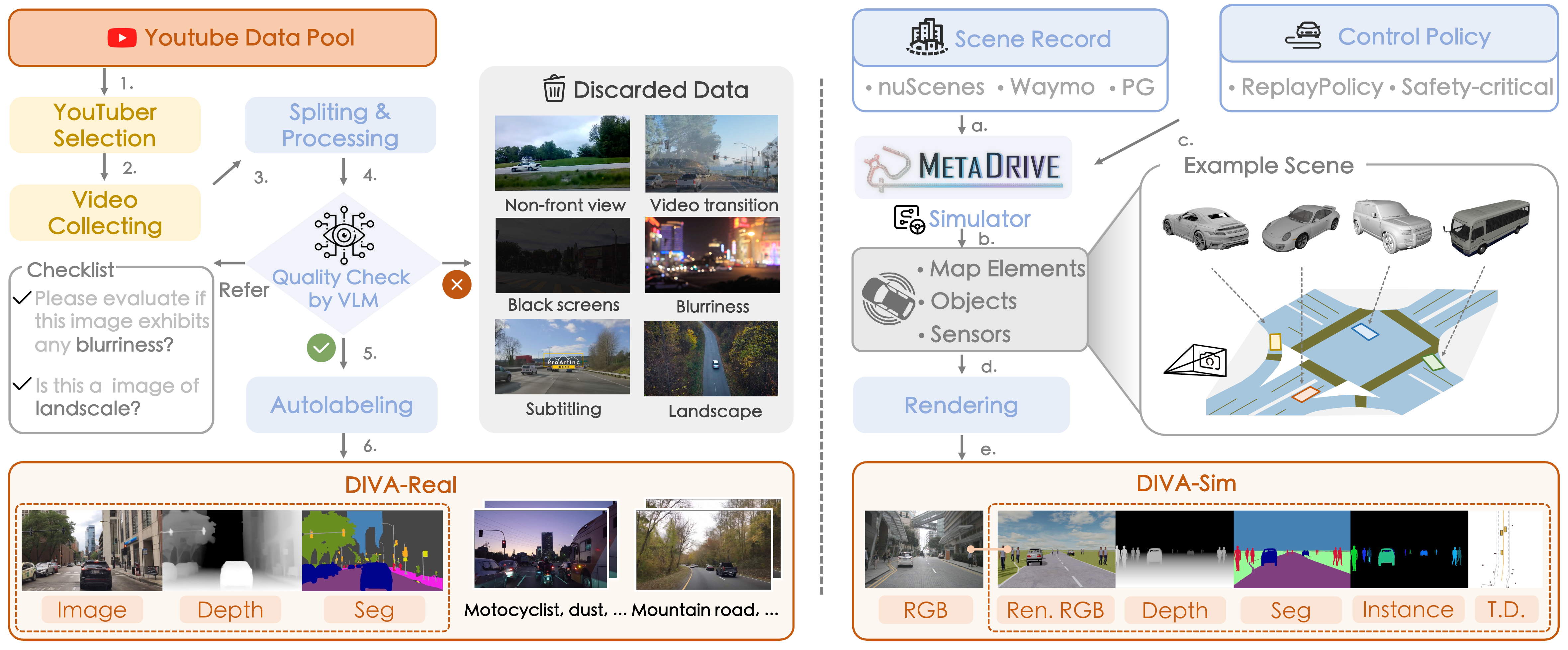}
    \caption{\textbf{Constructing DIVA dataset.}
    DIVA-Real (left) comprises driving videos collected from YouTube. We apply a Vision Language Model to filter out noisy images via a checklist and utilize off-the-shelf models to annotate text, depth, and semantic labels.
    Meanwhile, DIVA-Sim (right) employs scene records and control policies in a simulator to create map elements and objects.
    It can generate digital twins of real-world data and safety-critical scenes.
    Then various kinds of sensors placed in the simulation produce multimodal images. \texttt{Ren.}:rendered; \texttt{T.D.}: top-down view. Numbers and letters indicate the sequence of processes.
    }
    \label{fig:diva}
    \vspace{-7pt}
\end{figure}


%
%
Simulators are capable of faithfully reconstructing real-world scenes and hence obtaining training data with layout diversity. Also, after loading the driving scenarios such as map topology from the dataset, the simulator allows changing the motions and states of the traffic participants with pre-defined rules or interactive policies that differ from the original ones.
This inspires us to build \textbf{Sim2Real data} from the simulator. The Sim2Real data is induced from the same real-world scenarios, in which we can obtain real-world map topology, layout, and raw sensor data. At the same time, we can reconstruct the paired data from those scenarios but with reconstructed sensor data and even with altered layout and traffic flows.
DIVA-Sim utilizes the MetaDrive simulator~\cite{li2022metadrive} and ScenarioNet~\cite{li2023scenarionet} to gather 5.5 hours of virtual driving videos from nuScenes layouts~\cite{caesar2019nuscenes} and another 22 hours from procedurally generated behaviors.
%
%
%
It includes a set of safety-critical driving data through interactions introduced by an adversarial traffic generation method~\cite{zhang2023cat}, further improving the diversity of our dataset.

\noindent \textbf{Scene layout construction.}
%
%
%
We utilize ScenarioNet~\cite{li2023scenarionet} to transform scenes into a unified description format suitable for simulators, known as \textit{scene records}, logging map elements and objects.
As illustrated by the example scene in \cref{fig:diva} (right), loading \textit{scene records}, MetaDrive~\cite{li2022metadrive} can reconstruct roads, blocks, and intersections, and place corresponding 3D models like vehicles, cyclists, and pedestrians, based on the recorded positions and orientations.
We will reasonably select representative 3D models based on the category and dimensions of the objects.
And the model's shape is scaled based on the real dimensions to replicate the objects in the nuScenes dataset accurately.
By doing so, the digital twin scenario can be faithfully reconstructed in the simulator.

\noindent \textbf{Obtaining images via trajectory replay and rendering pipeline.}
%
The \textit{control policy} determines the motion dynamics, while the sensors generate multimodal image data at any desired location.
%
%
To create nuScenes digital twins, ReplayPolicy is applied to replay logged trajectories of all objects.
%
Our cameras are placed in the exact pose of the nuScenes front camera, with the camera's field of view adjusted to match that of nuScenes closely.
The camera attribute can be set to multiple types to obtain a variety of sensor data.
In summary, we can obtain the following conditions through the simulator: rendered RGB, depth, semantic segmentation, instance segmentation, and top-down views.

\noindent \textbf{Creation of safety-critical data.} 
Besides building digital twins of the real-world data, we can harness the simulator to continue growing the safety-critical data and enhance layout diversity. We apply the CAT method~\cite{zhang2023cat} to generate safety-critical data based on real-world scenarios. Specifically, 
we first randomly sample one scenario from the Waymo Open dataset~\cite{sun2020scalability}. 
%
A traffic vehicle is perturbed to attempt colliding with the ego-vehicle via adversarial interaction learning~\cite{zhang2023cat}. 
Thus, we harvest many safety-critical scenarios with adversarial driving behaviors, which might be challenging to collect in the real world.
This scalable creation of the safety-critical data from the simulator is also one of the strengths of our method. 
%
%


\vspace{-7pt}
\section{SimGen Framework}
\vspace{-5pt}

\begin{figure}[t]
    \centering
    \includegraphics[width=1\linewidth]{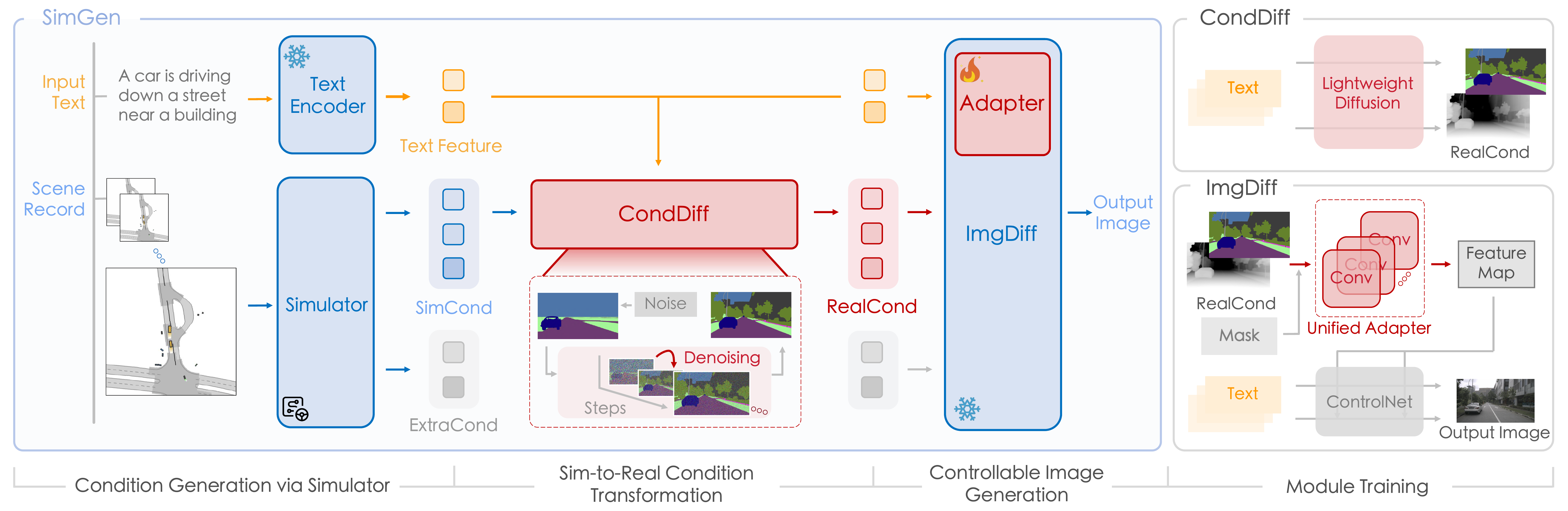}
    \caption{\textbf{Illustration of SimGen.}
    SimGen processes text and scene record as inputs. The text is feature-encoded and utilized in the subsequent modules, whereas the scene record undergoes a simulator rendering into simulated depth and segmentation (SimCond) and extra conditions (ExtraCond). 
    SimCond, coupled with the text features, is fed into the CondDiff module that converts SimCond into RealCond, representing real depth and segmentation. 
    Eventually, the text features, RealCond, and ExtraCond are inputted into the ImgDiff module, where an Adapter merges multi-source conditions into a unified control condition and generates driving scene images.
    }
    \vspace{-5pt}
    \label{fig:overview}
\end{figure}


SimGen aims to generate realistic driving images based on the text prompt and the spatial conditions including semantic and depth maps from real-world datasets and the driving simulator.
We incorporate a driving simulator into the data generation pipeline to achieve controllable and diverse image generation. Incorporating the simulator provides access to diverse layouts and behaviors of traffic participants, thus better closing the Sim2Real gap.
However, if just conditioning the diffusion model on synthesized data from a simulator, the diffusion model will result in bad image quality due to the limited assets and the artificial rendering.
We propose a cascade generative model that first transforms the simulated spatial conditions to realistic conditions as those in the dataset, then uses those realistic conditions to guide the first-view image diffusion model.

Illustrated in \cref{fig:overview}, SimGen first samples a driving scenario and a text prompt from the dataset and invokes the driving simulator MetaDrive~\cite{li2022metadrive} to render \textit{simulated conditions (SimCond)}, the synthesized depth and segmentation images.
%
%
Then, the SimCond and text features are fed into a lightweight diffusion model \textbf{CondDiff} (\cref{sec:CondFiff}) that converts simulated conditions into \textit{realistic conditions (RealCond)}, that resembles the real-world depth and segmentation images from YouTube and nuScenes datasets.
%
Finally, a diffusion model called \textbf{ImgDiff} (\cref{sec:ImgDiff}) generates a driving scene according to multi-modal conditions, including RealCond, textual prompts, and optional simulated spatial conditions, including RGB images, instance maps, and top-down views, \textit{etc}.

\subsection{Sim-to-Real Condition Transformation}
\label{sec:CondFiff}
While we strive to align the simulator settings with real data, such as intrinsic and extrinsic parameters of the camera, there is still a disparity between RealCond and SimCond.
The disparity arises from image mismatch, inherent flaws of the 3D models, and the simulator's lack of background details (\cref{sec:cascade}). 
%
%
Consequently, simulator conditions require transformation to closely resemble real ones.
An easy solution is to use domain adaptation \cite{mullick2023domain} and consider the SimCond and RealCond as different image styles.
However, training a domain transfer model that can generalize to novel scenarios requires paired SimCond and RealCond data far exceeding public datasets like nuScenes.
%
Thus, it's necessary to have an adaptation-free approach for Sim2Real transformation without additional training on SimCond.
%
To achieve that, we first use data from DIVA-Real to train a diffusion model, CondDiff, that generates RealCond purely from text prompts. 
The training does not contain data rendered from simulators. 
During inference, CondDiff injects noise-added SimCond into the intermediate sampling process and converts it into realistic conditions via continuous denoising.

%

\noindent \textbf{Learning to generate conditions from text inputs.} 
%
%
To facilitate the learning process of CondDiff, we initiate this stage with text-to-RealCond generation. 
Concretely, we utilize Stable Diffusion 2.1 (SD-2.1)~\cite{Rombach_2022_CVPR}, a large-scale latent diffusion model for text-to-image generation.
%
It is implemented as a denoising UNet, denoted by $\epsilon_{\theta}$, with multiple stacked convolutional and attention blocks, which learns to synthesize images by denoising latent noise.
Let $\mathbf{x}_0 \in \mathcal X$ represents a latent feature from the data distribution $p(\mathbf{x})$. Starting from $\mathbf{x}_0$, the training process involves gradually adding noise to procedure $\mathbf{x}_t$ for $t\in (0,1]$ until $\mathbf{x}_t$ transforms into Gaussian noise, namely forward stochastic differential equation (SDE)~\cite{ho2020denoising}.
The model is optimized by minimizing the mean-square error:
\begin{align}
\mathbf{x}_t=\alpha_t \mathbf{x}_0+\sigma_t &\epsilon, \epsilon \sim \mathcal{N}(\mathbf{0},\mathbf{I}), \mathbf{x}_0 \sim p(\mathbf{x}),
\\
\forall t,\ \underset{\theta}{\text{min}}\ \mathbb{E}||&\mathbf{\epsilon}-\epsilon_{\theta}(\mathbf{x}_t;\mathbf{c},t)||_2^2,
\label{eq:sd_train}
\end{align}
%
where $\sigma_t$ is a scalar function that describes the magnitude of the noise $\epsilon$ at denoising step $t$, $\alpha_t$ is a scalar function that denotes the magnitude of the data $\mathbf{x}_0$, $\theta$ parameterizes the denoiser model $\epsilon_{\theta}$, $\mathbf{\epsilon}$ is the added noise, and $\mathbf{c}$ is the text condition that guides the denoising process. 
The learning occurs in a compressed latent space $ \mathcal X$ instead of the pixel space \cite{Rombach_2022_CVPR}. 
During sampling, the model iteratively denoises the final step prediction from the standard Gaussian noise to generate images.

The original SD-2.1 is trained on data from various domains unrelated to the depth and semantic images in driving scenes. 
As depicted in the CondDiff in the upper right of \cref{fig:overview}, we fine-tune the SD-2.1 to be a text-to-RealCond model using the triplets of text, depth and segmentation data from DIVA-Real and nuScenes,
with the objective of \cref{eq:sd_train}.
After loading the SD-2.1 checkpoint, all parameters $\theta$ of the UNet are fine-tuned at this stage, while the CLIP text encoder \cite{radford2021learning} and autoencoder \cite{esser2021taming} remain frozen.
%
The depth and segmentation data is autolabelled by a set of perception models as discussed in \cref{section:diva-real}.
%

\noindent \textbf{Adaptation-free sim-to-real transformation.} 
Now, we have a model CondDiff that generates RealCond purely from text prompts. 
We will then use the conditions from simulator \textit{SimCond} to guide the sampling process so that we can transform SimCond to RealCond.
%
According to SDEdit~\cite{meng2021sdedit}, the reverse SDE, where the diffusion model iteratively denoises standard Gaussian noise to generate images, can start from any intermediate time.
This inspires us to insert noise-added SimCond into the intermediate time of the sampling process, and the model will use them as guidance to generate RealCond with the SimCond layouts.
%
%
%
%
In detail, the module first encodes the SimCond into latent space to get $\mathbf{x}^{\text{sim}}$. It selects a specific
time $t_s\in (0,1)$ and perturbs the input $\mathbf{x}^{\text{sim}}$ using a Gaussian noise of standard deviation $\sigma^2_{t_s}$ as follows:
\begin{equation}
    \texttt{Sample}\ \mathbf{x}^{\text{noi}}\sim \mathcal{N}(\mathbf{x}^{\text{sim}}; \sigma^2_{t_s}\mathbf{I}).
\end{equation}
The perturbing process will effectively remove low-level details like pixel information while preserving high-level cues like rough color strokes~\cite{meng2021sdedit}.
The noise-processed image $\mathbf{x}^{\text{noi}}$ seamlessly substitutes the diffusion model's state at time $t_s$ during denoising.
Thus, the intermediate state $\mathbf{x}_{t_s}=\mathbf{x}^{\text{noi}}$ serves as a guidance to solve the corresponding reverse SDE as follows:
\begin{equation}
    p_{\theta}(\mathbf{x}_{t_s-1}|\mathbf{x}_{t_s}) = \mathcal{N}(\mathbf{x}_{t_s-1};\mu_{\theta}(\mathbf{x}_{t_s},t),\Sigma_\theta(\mathbf{x}_{t_s},t_s)),
\end{equation}
where $\mu_{\theta}$ and $\Sigma_{\theta}$ are determined by CondDiff $\epsilon_{\theta}$.
The above equation iterates until the model generates a synthesized image $\mathbf{x}_0$ like RealCond at $t_s=0$.
Throughout this process, all parameters of CondDiff remain frozen, with only SimCond $\mathbf{x}^{\text{sim}}$, text $\mathbf{c}$, and noise affecting the sampling process.
%


\subsection{Controllable Image Generation with Multimodal Conditions}
\label{sec:ImgDiff}



\begin{wrapfigure}{r}{0.36\textwidth}

\vspace{-12pt}
\captionof{table}{\textbf{Formats of conditions.}
\texttt{Real/SimCond}: depth and segmentation;
\texttt{ExtraCond}: rendered RGB, instance maps, and top-down views.
}
\vspace{-5pt}
\centering
\scriptsize
\tablestyle{2pt}{1.05}
\setlength{\tabcolsep}{1mm}{
\begin{tabular}{lccc}
\toprule
\scriptsize{Dataset}    & \scriptsize{RealCond} & \scriptsize{SimCond}  & \scriptsize{ExtraCond} \\ \midrule
\scriptsize{nuScenes}   & \scriptsize{\ding{51}}  &   &   \\
\scriptsize{\baseline{DIVA-Real}}   & \baseline{\scriptsize{\ding{51}}}  & \baseline{}   & \baseline{} \\
\scriptsize{\baseline{DIVA-Sim}}   &  \baseline{} &    \baseline{\scriptsize{\ding{51}}} &  \baseline{\scriptsize{\ding{51}}}   \\
\bottomrule
\end{tabular}
}
\vspace{-5pt}
\label{tab:format}
\end{wrapfigure}

In the second stage, we will use a diffusion-based model to synthesize diverse driving images by integrating various control conditions (\cref{tab:format}), including the RealCond from the data or generated from SimCond by CondDiff, the textual prompt, and some extra conditions ExtraCond such as rendered RGB, instance segmentation, and top-down views from the simulator.
ExtraCond offers additional information for the output image, including road typology and object attributes (orientation, outlines, and 3D locations), highlighting the necessity of incorporating them into model control.
%
%

However, there exist conflicts among multimodal conditions (\cref{sec:adapter}):
1) \textit{Modal discrepancy}: The nuScene dataset contains a full set of RealCond, SimCond, and ExtraCond, while YouTube only includes RealCond. 
This might impact the quality of images generated based on nuScenes layouts due to the data bias for diffusion models \cite{kim2024training}.
2) \textit{Condition disparity}: The lack of rich background information in simulated conditions compared to real ones results in a struggle between the two modalities. 
In real-world images, the background might contain urban buildings with drastically different facades and street trees of different species. 
Although CondDiff can convert SimCond to RealCond, the domain gap prevents the same transformation for ExtraCond (\textit{e.g.}, rendered RGB, instance segmentation, and top-down views) from the simulator.
%
%
Thus, we propose using a unified adapter in ImgDiff to address these issues.
Its essence lies in mapping variable conditions into fixed-length vectors, overcoming the misalignment of low-level features, and enabling a unified control input interface for the diffusion model.

\noindent
\textbf{Mitigating condition conflicts with adapters.}
Adapters are essential at the guiding branch of image generation to ensure the model learns necessary, unique, non-conflicting information from all conditions.
Inspired by UniControl~\cite{qin2023unicontrol}, we devised a set of convolutional modules as the adapters to capture features from various modalities, as shown in the ImgDiff in the lower right of \cref{fig:overview}. 
For a set of input conditions $\mathbf{X}=\{\mathbf{x}^1, \mathbf{x}^2, ..., \mathbf{x}^K\}$, each condition undergoes feature extraction via the unified adapter $\mathcal{F}_\text{ada}$ represented as:
\begin{equation}
    \mathcal{F}_{\text{ada}}(\mathbf{x}^{k}):= \sum^{K}_{i=1}\mathbbm{1}_{i=k}  \mathcal{F}^{(i)}_{\text{cov1}}
    (
        \mathcal{F}^{(i)}_{\text{cov2}}(\mathbf{x}^{k}\cdot \mathbf{M}^{k})
    )    
    ,   
\end{equation}
where $\mathbbm{1}$ is the indicator function, $\mathbf{x}^{k}$ is the $k$-th condition image, and $\mathcal{F}^{(i)}_{\text{cov1}}$, $\mathcal{F}^{(i)}_{\text{cov2}}$ are the convolution layers of the $i$-th module of the adapter.
$\mathbf{M}^{k}$ is the valid mask for each condition.
%

The valid mask is the key to mitigating conflicts.
The entire mask will be padded with 0 if a condition is missing or not provided.
For simulator-generated conditions, we set the masks of backgrounds to 0 based on the semantic labels, preventing unwanted constraints on background generation.
Since top-down view conditions don't belong to the frontal perspective, all information is retained.
Ultimately, two convolutional layers process the concatenated condition features, max pooling them into a fixed-length feature vector for control.



\noindent
\textbf{Controllable image generation.}
We utilize the ControlNet~\cite{zhang2023adding} to guide image generation. 
After the feature extraction by $\mathcal{F}_\text{ada}$, conditions are encoded into the UNet model.
%
Then, the model injects control information into each UNet layer through residual connections.
All parameters in UNet's input and middle layers are frozen, and we only fine-tune the output layers and the control branch.

%


\vspace{-3pt}
\section{Experiments}
\vspace{-3pt}

\textbf{Setup and protocols.}
SimGen is learned in two stages on DIVA and nuScenes dataset~\cite{caesar2019nuscenes}. 
%
The performance is evaluated based on image quality, controllability, and diversity. The Frame-wise Fréchet Inception Distance (FID) evaluates the synthesized data's quality. SimGen's controllability corresponds to how well the generated images align with ground truths from the nuScenes validation set. The controllability is measured by the 3D detection metrics (AP) and BEV segmentation metrics (mIoU) when applying out-of-the-box perception models on the generated images. Lastly, diversity is measured using the pixel variance of the generated images.
%
%
%
More details on training, sampling, and evaluation metrics are provided in \cref{sec:implementation} and \cref{sec:supp_exp}.

\subsection{Comparison to State-of-the-arts}

\begin{wrapfigure}{r}{0.5\textwidth}
\vspace{-12pt}
\captionof{table}{\textbf{Generation quality and diversity compared to nuScenes experts.}
The FID and $D_{\text{pix}}$ indicate the image quality and pixel diversity, respectively.
\colorbox{baselinecolor}{gray}: main metric.
\textbf{bold}: best results.
}
\vspace{-3pt}
\centering
\footnotesize
\tablestyle{2.0pt}{1.05}
\setlength{\tabcolsep}{1mm}{
\begin{tabular}{lccc}
\toprule
Method    & Dataset & FID$\downarrow$  & \baseline{$D_{\text{pix}}\uparrow$}  \\
\midrule
BEVGen~\cite{swerdlow2024street}   & \multirow{5}{*}{nuScenes}  &  25.5 &  \baseline{17.0} \\
BEVControl~\cite{yang2023bevcontrol}   &   &   24.9 &  \baseline{-} \\
MagicDrive~\cite{gao2023magicdrive}   &   &             16.6 &  \baseline{19.7}   \\
Panacea~\cite{wen2023panacea} &    & 17.0 &  \baseline{-}     \\
DrivingDiffusion~\cite{li2023drivingdiffusion} &      &    15.9 & \baseline{20.1}  \\ \midrule
SimGen-nuSc   & nuScenes    &  \textbf{15.6}   & \baseline{20.5} \\
\textbf{SimGen}   & \textbf{DIVA}  &  \textbf{15.6}   & \baseline{\textbf{26.6}}\\
\bottomrule
\end{tabular}
}
\label{tab:quality}
\end{wrapfigure}

\begin{figure}[t]
    \centering
    \includegraphics[width=0.99\linewidth]{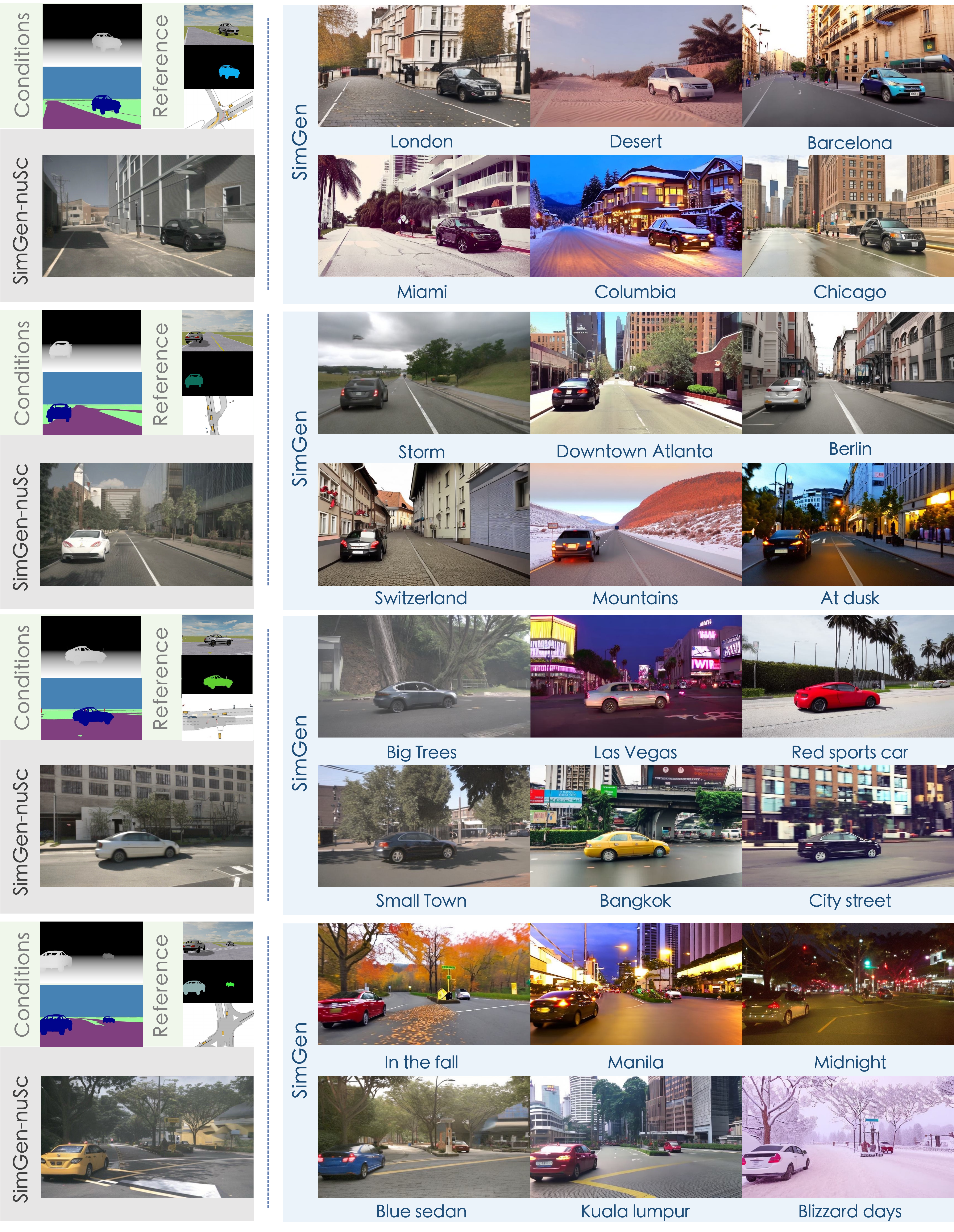}
    \caption{\textbf{Generating diverse appearances conditioned on simulator's conditions and texts.}
    We show the generation results of SimGen (blue boxes) and SimGen-nuSc (gray boxes) under the same conditions.
    Compared to models confined to limited datasets, SimGen exhibits a stronger ability to generate more realistic and diverse driving scenarios.
    Reference is not used for inference.
    }
    \vspace{-15pt}
    \label{fig:main_vis}
\end{figure}

\textbf{Comparison to nuScenes-specific models.}
We compare SimGen with the most recently available data generation approaches exclusively trained on nuScenes.
\cref{tab:quality} shows that SimGen surpasses all previous methods in image quality (FID) and diversity ($D_{\text{pix}}$).
Specifically, SimGen significantly increases $D_{\text{pix}}$ by \textbf{+6.5} compared to DrivingDiffusion \cite{li2023drivingdiffusion}. 
For fair comparisons, we train a model variant (SimGen-nuSc) on the nuScenes dataset only. 
We find that although SimGen-nuSc performs on par with SimGen on nuScenes, its performance in diversity is less than ideal, and it struggles to generalize to novel appearances like \texttt{Desert}, \texttt{Mountains}, and \texttt{Blizzard}, where the generation degrades to the nuScenes visual pattern. 
In contrast, SimGen trained on DIVA exhibits strong generalization ability across appearances as shown in \cref{fig:main_vis}.


\textbf{Controllability for autonomous driving.}
The controllability of our method is quantitatively assessed based on the perception performance metrics obtained using a single-frame version of BEVFusion~\cite{liang2022bevfusion}.
We feed the data from nuScenes validation set into SimGen and generate the driving images.
Then, the perception performance of pre-trained BEVFusion, involving map segmentation (mIoU) and 3D object detection (AP), is recorded.
Compared to the perception scores on the raw nuScenes data, the relative performance metrics serve as the indicators of the alignment between the generated images and the conditions.
As depicted in \cref{tab:control}, SimGen achieves a relative performance of \textbf{-3.3} on map segmentation of vehicles, underscoring a robust alignment of the generated samples.

\textbf{Data augmentation via synthetic data.}
%
%
SimGen can produce augmented data with accurate annotation controls, enhancing the training for perception tasks, \textit{e.g.}, map segmentation, and 3D object detection.
For these tasks, we augment an equal number of images as in nuScenes dataset, ensuring consistent training iterations and batch sizes for fair comparisons to the baseline.
%
\cref{tab:aug} indicates that blending generated with real data can elevate the singe-frame version of BEVFusion's vehicle mIoU to \textbf{39.0}, a \textbf{+4.4} uptick compared to models trained purely on real data. 
These outcomes reinforce SimGen's validity as a controllable synthetic data generator for enhancing perception models.

\begin{table}[t]
\begin{minipage}[t]{0.54\linewidth}
\makeatletter\def\@captype{table}\makeatother\caption{\textbf{Generation controllability for perception tasks.}
\texttt{Oracle}: a single-frame version of BEVFusion~\cite{liang2022bevfusion}.
In blue is the relative drop compared to standard nuScenes validation
data. 
}
\vspace{5pt}
\centering
\footnotesize
\tablestyle{2.0pt}{1.05}
\setlength{\tabcolsep}{1mm}{
\begin{tabular}{lcccc}
\toprule
\multirow{2}{*}{Method} & \multicolumn{2}{c}{Map Seg} & \multicolumn{2}{c}{Object Detection} \\
\cmidrule(r){2-5}
& \scriptsize{$\text{mIoU}_{\text{Road}}$}   & \scriptsize{\baseline{$\text{mIoU}_{\text{Vehicle}}$}}   & \scriptsize{$\text{AP}_{\text{Car}}$}    & \scriptsize{$\text{AP}_{\text{Truck}}$}     \\ \midrule
\color{gray}{Oracle}    & \color{gray}{72.2}     & \color{gray}{\baseline{34.6}}     & \color{gray}{47.0}     & \color{gray}{21.4}     \\
BEVGen~\cite{swerdlow2024street}  & 50.1 \tiny\textcolor{DarkBlue}{(-21.1)}  & \baseline{5.9} \tiny\textcolor{DarkBlue}{(-28.7)}     & 24.7 \tiny\textcolor{DarkBlue}{(-22.3)}  & 9.1 \tiny\textcolor{DarkBlue}{(-15.0)}   \\
MagicD.~\cite{gao2023magicdrive}   & 58.6 \tiny\textcolor{DarkBlue}{(-13.6)}   & \baseline{29.5 \tiny\textcolor{DarkBlue}{(-5.1)}}    & 37.3 \tiny\textcolor{DarkBlue}{(-9.7)}   & 17.3 \tiny\textcolor{DarkBlue}{(-4.1)}  \\ \midrule
SimGen-nuSc  & 60.6 \tiny\textcolor{DarkBlue}{(-11.6)}   & \baseline{29.9 \tiny\textcolor{DarkBlue}{(-4.7)}}     & 39.1 \tiny\textcolor{DarkBlue}{(-7.9)}    & 18.1 \tiny\textcolor{DarkBlue}{(-3.3)} \\
\textbf{SimGen}  & \textbf{62.9 \tiny\textcolor{DarkBlue}{(-9.3)}}   & \baseline{\textbf{31.2 \tiny\textcolor{DarkBlue}{(-3.4)}}}     & \textbf{41.0 \tiny\textcolor{DarkBlue}{(-6.0)}}     & \textbf{19.6 \tiny\textcolor{DarkBlue}{(-1.8)}}       \\ \bottomrule
\end{tabular}
}
\label{tab:control}
\end{minipage}
\hfill
\begin{minipage}[t]{0.44\linewidth}
\makeatletter\def\@captype{table}\makeatother\caption{
\textbf{Comparison involving data augmentation using synthetic data.}
The \texttt{Baseline} is a single-frame version of BEVFusion~\cite{liang2022bevfusion} trained on nuScenes train set.
}
\vspace{5pt}
\centering
\footnotesize
\tablestyle{2.0pt}{1.05}
\setlength{\tabcolsep}{1mm}{
\begin{tabular}{lcccc}
\toprule
\multirow{2}{*}{Method} & \multicolumn{2}{c}{Map Seg} & \multicolumn{2}{c}{Object Det} \\
\cmidrule(r){2-5}
& \scriptsize{$\text{mIoU}_{\text{Road}}$}   & \scriptsize{\baseline{$\text{mIoU}_{\text{Vehi}}$}}   & \scriptsize{$\text{AP}_{\text{Car}}$}    & \scriptsize{$\text{AP}_{\text{Truck}}$}     \\ \midrule
Baseline    & 72.2     & \baseline{34.6}     & 47.0     & 21.4     \\
BEVGen~\cite{swerdlow2024street}  & 71.9   & \baseline{34.2}      & 47.3   & 21.1    \\
MagicD.~\cite{gao2023magicdrive}   & 77.4    & \baseline{37.7}    & 48.0    & 22.8   \\ \midrule
SimGen-nuSc  & 77.7   & \baseline{38.0}     & 48.3     & 23.0       \\
\textbf{SimGen}  & \textbf{78.9}   & \baseline{\textbf{39.0}}     & \textbf{49.1}     & \textbf{23.6}       \\ \bottomrule
\end{tabular}
}
\label{tab:aug}
\end{minipage}
\vspace{-10pt}
\end{table}

\subsection{Ablation Study}
\vspace{-3pt}

\begin{wrapfigure}{r}{0.35\textwidth}
\begin{minipage}[b]{0.35\textwidth}
\footnotesize
\vspace{-12pt}
\captionof{table}{\textbf{Ablation on designs in SimGen.} All proposed
designs contribute to the final performance.}
\centering
\tablestyle{2.0pt}{1.05}
\setlength{\tabcolsep}{1mm}{
\begin{tabular}{lcc}
\toprule
Ablation   & FID$\downarrow$ & $\text{AP}_{\text{Car}}\uparrow$ \\ \midrule
Baseline   & 19.5   & 45.7          \\
+ Cascade Pipeline & 17.2 & 46.3 \\
+ ExtraCond &  17.7  & 47.6          \\
+ Unified Adapter  & \textbf{16.9}   & 48.2          \\ \bottomrule
\end{tabular}
}
\label{tab:ablation}
\vspace{-5pt}
\end{minipage}
\end{wrapfigure}

\begin{figure}[t]
    \centering
    \includegraphics[width=0.99\linewidth]{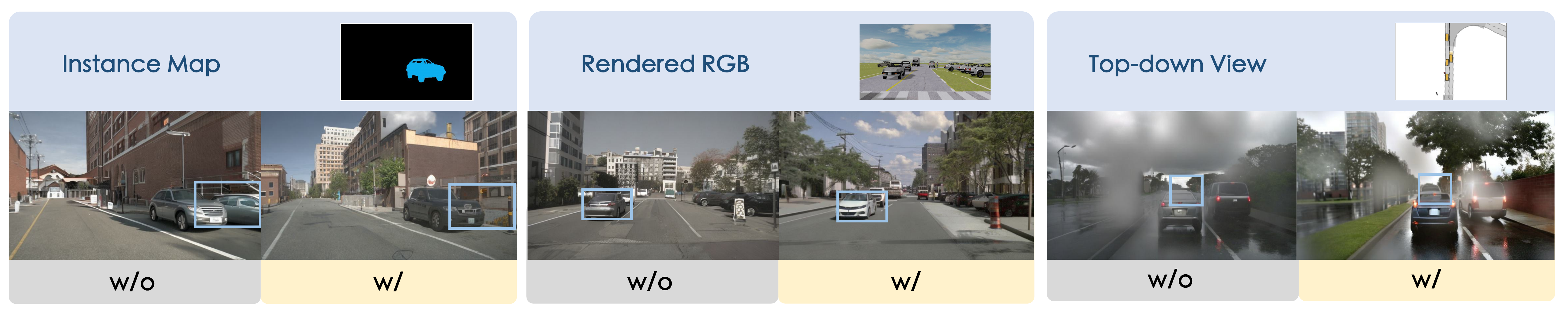}
    \caption{\textbf{Ablation study of simulator conditions.}
}
\label{fig:ablation_vis}
\vspace{-20pt}
\end{figure}

The ablation is conducted by training each variant of our model on a DIVA subset with 30K frames, and we report FID and average precision of cars ($\text{AP}_{\text{Car}}$) as the quality and controllability metrics.
We gradually introduce our proposed components and conditions, starting with a ControlNet baseline~\cite{zhang2023adding} that directly takes SimCond as input.
As shown in \cref{tab:ablation}, by introducing a cascade pipeline to
transform SimCond into RealCond, the FID significantly reduces by \textbf{-2.3}, as the transformed conditions closely resemble real scenarios. 
Including simulator-pulled ExtraCond to the control conditions improves the alignment of the generated images with the target layouts, effectively enhancing the $\text{AP}_{\text{car}}$ by \textbf{+1.3}. 
However, a slight deterioration in the FID metric (\textbf{+0.5}) may result from condition conflicts. 
Lastly, using a Unified Adapter helps alleviate conflicts, significantly improving generated image quality by \textbf{-0.8}. 
The effectiveness of ExtraCond is exhibited in \cref{fig:ablation_vis}, where the addition of instance map, rendered RGB, and top-down view enables the model to better handle object boundaries, orientation angles, and occlusions in these cases.
To the best of image quality, we only use depth and segmentation conditions in subsequent experiments.

\begin{figure}[t]
    \centering
    \includegraphics[width=0.99\linewidth]{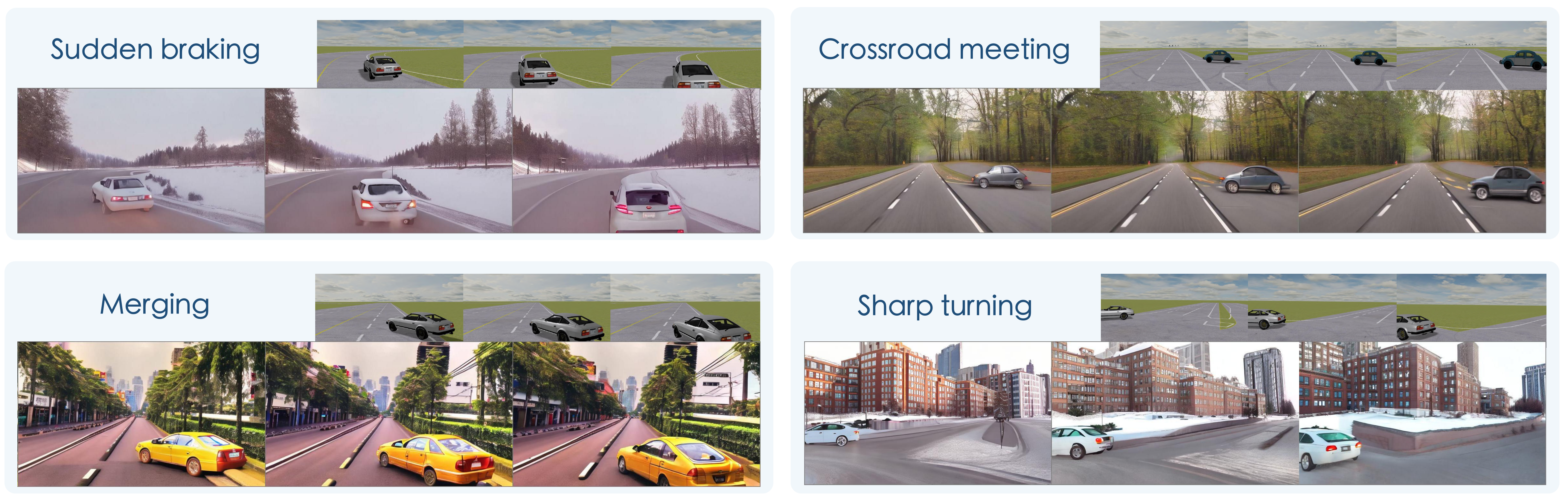}
    \caption{\textbf{Generating safety-critical scenes.}
SimGen can also recreate image sequences of safety-critical scenes where risky driving behaviors like sudden braking and merging happen.  }
\label{fig:layout_vis}
\vspace{-10pt}
\end{figure}

\vspace{-3pt}
\subsection{Discussions}
\vspace{-3pt}

\noindent
\textbf{Extension to video generation.}
SimGen is not designed for video generation. But the high-quality image generation brings a potential for
video generation, which is important for interactive scene generation and closed-loop planning.
We have a preliminary attempt by integrating temporal attention layers into UNet similar to \cite{yang2024generalized}, and then conducting subsequent training stages focusing solely on learning the newly added layers while freezing the original parameters. 
This shows a promising result of temporal consistency across frames, as compared with video generation models in \cref{sec:more_ablation}.

\textbf{Generating safety-critical scenarios.}
The key innovation of SimGen is the controllability of layouts brought by connecting to a driving simulator.
%
Building upon video generation, we showcase SimGen's generalization capabilities in novel layouts, specifically in safety-critical scenarios in \cref{fig:layout_vis}.
The visualized layout is initialized from a scenario sampled from the Waymo Open dataset~\cite{sun2020scalability} and then populated with risky behaviors via an adversarial interaction traffic flow generation method~\cite{zhang2023cat}.
%
%
SimGen can transform safety-critical driving scenarios from the simulator into realistic sequential images, including risky behaviors like \texttt{sudden braking}, \texttt{crossroad meeting}, \texttt{merging}, \texttt{sharp turning}, \textit{etc}.
This application is impossible with existing models, which are only trained and conditioned on a given static real-world dataset that lacks records of dangerous driving behaviors.
%
This brings new opportunities for closed-loop data generation capabilities (\cref{sec:more_ablation}).

\vspace{-3pt}
\section{Related Work}
\vspace{-3pt}

%

\noindent
\textbf{Diffusion-based generative Models.}
Diffusion models have made significant strides in image generation~\cite{dhariwal2021diffusion,meng2021sdedit,nichol2021glide,podell2023sdxl,ramesh2022hierarchical,saharia2022photorealistic} and video generation~\cite{blattmann2023stable,he2022lvdm}. Recent works incorporate additional control signals beyond text prompts~\cite{guo2023animatediff,li2023gligen,mou2024t2i}. 
ControlNet~\cite{zhang2023adding}
integrates a trainable copy of the SD encoder for control signals. 
Studies like Uni-ControlNet~\cite{zhao2024uni} and UniControl~\cite{qin2023unicontrol} have also focused on fusing multimodal inputs into a unified control condition using input-level adapter structures. 
Our method distinguishes itself in its capability of multimodal conditioned generation by addressing the sim-to-real gap and condition conflicts in the complex realm of driving scenarios.

\noindent
\textbf{Controllable generation for autonomous driving.}
Autonomous driving research heavily relies on paired data and layout ground truths, spurring numerous studies on their generation~\cite{cheng2023layoutdiffuse, lu2023wovogen}. 
Some works~\cite{gao2024vista,hu2023gaia,yang2024generalized} utilize diffusion models to generate future driving scenes based on historical information, but they lack the ability to control scenes through layout.
Other generative methods, like BEVGen~\cite{swerdlow2024street} and BEVControl~\cite{yang2023bevcontrol}, use BEV layouts to create synthetic single or multi-view images. 
Recent innovative method Panacea~\cite{wen2023panacea} generates panoramic and controllable videos, while MagicDrive~\cite{gao2023magicdrive} offers diverse 3D controls and tailored encoding strategies. Lastly, DriveDreamer~\cite{wang2023drivedreamer} and DrivingDiffusion~\cite{li2023drivingdiffusion} employ diffusion models for realistic multi-view video generation and environment representation. 
Yet,
these works are confined to limited appearances and layouts of static datasets, restraining their real-world applicability and the controllability over the layouts that deviate from the dataset, such as the safety-critical scenarios.

\noindent
\textbf{Scenario generation via simulators.}
Driving simulators~\cite{dosovitskiy2017carla,li2022metadrive} are fundamental to autonomous driving development, providing controlled simulations that mimic reality. 
Notable studies include SYNTHIA~\cite{ros2016synthia}, AIODrive~\cite{weng2023all}, and GTA-V~\cite{richter2016playing} that generate vitural images and annotations.
SHIFT~\cite{sun2022shift} diversifies with environmental changes, while CAT~\cite{zhang2023cat} creates safety-critical scenarios for targeted training from real-world logs. 
Despite their layout diversity and attempts at photorealism enhancement~\cite{richter2022enhancing}, the simulated images lack realism. 
In this work, we bridge the two worlds to obtain both the appearance diversity from diffusion models and the layout controllability from simulators.


\vspace{-3pt}
\section{Conclusion}
\label{sec:conclusion}
\vspace{-3pt}

We propose a simulator-conditioned diffusion model, SimGen, that learns to generate diverse driving scenarios by mixing data from the simulator and the real world.
A novel dataset containing massive web and simulated driving videos is collected to ensure diverse scene generation and mitigate simulation-to-reality gap.
By obtaining diversity in appearance and layout, SimGen exhibits superior data augmentation and zero-shot generalization capabilities in generating diverse and novel scenes.



\noindent
\textbf{Limitations and future work.}
SimGen currently does not support multi-view generation, limiting its application in Bird's Eye View models.
Inheriting the drawbacks of diffusion models, SimGen suffers from long inference time,
which may impact the applications like closed-loop training. The study of extending SimGen to video generation is left for future work.

\vspace{-5pt}
\section*{Acknowledgements}
\vspace{-5pt}

The project was supported by the NSF Grants CCRI-2235012 and RI-2339769, and the Sony Focused Research Award. YZ, HZ, and MG were supported by the National Natural Science Foundation of China (No. 62432008).
ZP is supported by the Amazon Fellowship via the Science Hub for Humanity and Artificial Intelligence at UCLA.

{\small
\bibliographystyle{ieeenat_fullname}
\bibliography{main,short}
}



\newpage

\noindent\textbf{\large{Appendix}}

\appendix

\startcontents
{
    \hypersetup{linkcolor=black}
    \printcontents{}{1}{}
}
\newpage

\section{Discussions}
\label{sec:discussions}

\textit{SimGen project page} provides links to the YouTube videos (\texttt{DIVA-Real\_Video\_Links}) used in DIVA-Real, as well as digital twins of nuScenes dataset (\texttt{DIVA-Sim\_nuSc\_Digital\_Twins}) and safety-critical video clips (\texttt{DIVA-Sim\_Safety-critical\_Demo\_Videos}) included in DIVA-Sim.

To better understand our work, we supplement with the following question-answering.

\textbf{Q1.}
\textit{What makes SimGen stand out compared to pixel-to-pixel transformation models?}

Recent GAN-based and Diffusion-based works in image transformation can generate images that are controllable based on specific conditions~\cite{guo2023animatediff,saharia2022photorealistic}.
Yet, their limitations lie in the fact that the content they generate is \textit{strictly} tethered to these input conditions.
If these conditions, derived directly from a simulator, are missing important contextual information like backgrounds and buildings, then output images may similarly lack background details
%
Consequently, SimGen employs a cascade structure, permitting the CondDiff model to conceptualize different background scenarios through text, thereby enriching the visual composition of the rendered driving scenes.
Detailed analysis is shown in \cref{sec:empirical study}.

\textbf{Q2.}
\textit{What is the criteria to demonstrate good generalization and diversity of your model? How much data do we need?}

Currently, it's challenging to define a specific standard to assess the diversity and generalization abilities of the models, as quality evaluation is subjective and fair comparison can be difficult. 
However, by utilizing publicly available data, we have found that scaling up the data size proves beneficial for the zero-shot generation on novel scenarios. 
Equally important to note is that our approach is easily scalable, and by leveraging massive in-the-wild data, we offer a continuing opportunity to strengthen its generalization capabilities.

\textbf{Q3.}
\textit{What is the definition of safety-critical scenarios and how to ensure they are realistic and feasible?}

A safety-critical scenario is a situation where one or more vehicles collide with the ego vehicle, which is rare to collect in real-world datasets like Waymo. We utilize CAT~\cite{zhang2023cat}  to generate risky behaviors from logged scenarios to ensure reality and feasibility, which uses a data-driven motion prediction model that predicts several modes of possible trajectories of each traffic vehicle. Please refer back to \cite{zhang2023cat} for a detailed description of safety-critical scenarios.



\textbf{Q4. Broader impact.}
\textit{What are potential applications and future directions with the provided DIVA data and the SimGen model, for both academia and industry?}

\textbf{Datasets.}
DIVA collects massive data from YouTube and simulators, significantly enhancing the appearance and layout diversity of driving video clips. 
This provides the community with extensive high-quality resources for exploring open avenues in autonomous driving and Sim2Real research. 

\textbf{Models.}
Beyond data augmentation, we hope our model can also benefit the community by enabling wider applications. 
In this work, we demonstrate SimGen's capability as a closed-loop data generator. 
It holds promise to adapt to downstream tasks like closed-loop evaluation of autonomous driving agents~\cite{peng2022reward}, which is showcased in \cref{sec:more_ablation}.
To boost deployment efficiency, distilling knowledge from the generative model is worth exploring~\cite{li2023dreamteacher}. 
Besides, simulator-conditioned scene generation also provides opportunities to achieve physically grounded real-world generation~\cite{ajay2024compositional,du2024learning}. 
Please note that our model will be publicly released to benefit the community and can be further fine-tuned flexibly according to custom data within the industry.

\textbf{Negative societal impacts.}
The potential downside of SimGen could be its unintended use in generating counterfeit driving scenarios. 
Our code includes the Diffusers~\cite{von2022diffusers} safety checker to screen for NSFW outputs.
Besides, we plan to regulate the effective use of the model and mitigate possible societal impacts through gated model releases and monitoring mechanisms for misuse.


\textbf{Q5. Limitations.}
\textit{What are the issues with current designs, and corresponding preliminary solutions?}

1)
Panoramic image generation is necessary for current Bird's Eye View perception models in autonomous driving.
Yet, to utilize scaled web data, which consists of front-facing single-camera footage, SimGen does not engage in multi-view image generation. 
This may limit the application of SimGen in real-world deployments.
2)
We chose SD-2.1~\cite{Rombach_2022_CVPR} as our base diffusion model, inheriting its advantages of high visual quality and better rendering capabilities of the text encoder. On the other hand, we noted that it has a slow sampling speed and high computation costs. Our model does indeed suffer from this issue.

However, as a pioneering work exploring how to introduce simulators into generative models for diverse driving scene generation, the primary focus of this work is the simulator-conditioned generalization ability across unseen driving scenarios rather than multi-view designs and computational overhead. Future work might include trying cross-frame attentions~\cite{gao2023magicdrive,li2023drivingdiffusion,wen2023panacea,deng2024streetscapes}, faster sampling methods~\cite{chen2024diffusion, meng2023distillation,salimans2022progressive,zhao2024unipc}, and transferring our general method to more efficient diffusion models.

\section{DIVA Dataset}
\label{sec:sup_diva}

Our dataset, DIVA, contains 147.5 hours of driving video along with diverse multimodal conditions, including text, segmentation, depth, and virtual images. 
In this section, we detail the YouTube and simulator video collection process, annotation method, more examples, and analysis to illustrate the diversity of the DIVA dataset.

\subsection{DIVA-Real}

\subsubsection{Data Collection and Cleaning}

\begin{figure}
    \centering
    \includegraphics[width=1\linewidth]{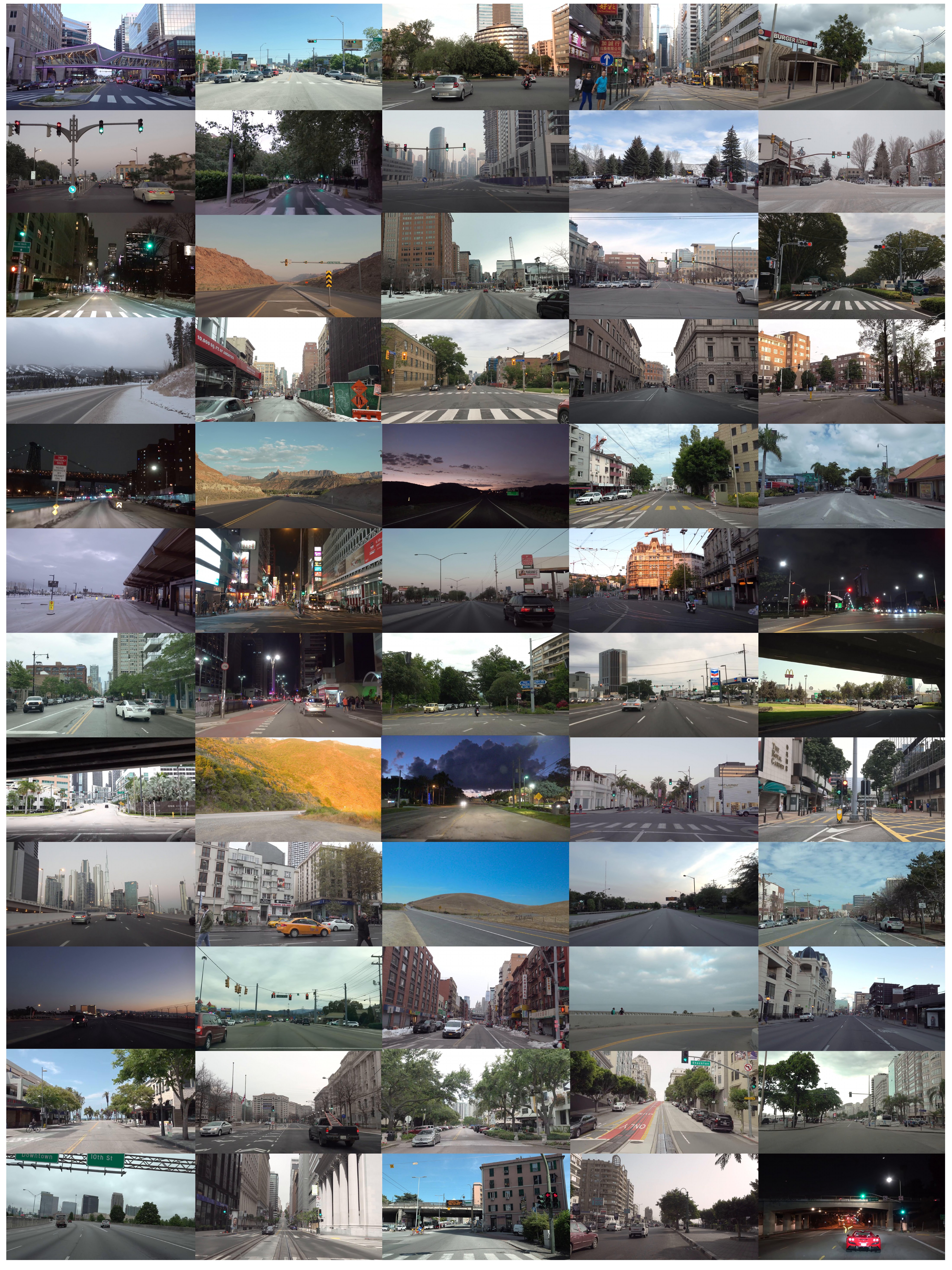}
    \caption{\textbf{Various video samples from DIVA-Real.} Due to space limitations, we only showcase certain frames from the videos. It covers a wide range of diversity across multiple axes, including geographical location, traffic scenarios, time periods, weather conditions, \textit{etc}.}
    \label{fig:diva_vis}
\end{figure}

\noindent
\textbf{Data preparation.}
We first searched for driving videos on YouTube using keywords such as \texttt{driving videos}, \texttt{4K}, and \texttt{HD}. 
We then identified a selection of YouTubers who consistently upload high-quality driving videos. 
We further inspected the quality of these videos in terms of resolution, resulting in 130 high-quality front-view driving videos, including \texttt{Barcelona 4K - Driving Downtown}, \texttt{Cairo 4K - Pyramid Expressway Sunrise}, \texttt{Las Vegas 4K - Sunset Drive}
and \texttt{Istanbul 4K - Night Drive - Turkey}. 
We used videos from 10 selected clips as the validation set and the other videos for training. The diversity of DIVA-Real is illustrated in \cref{fig:diva_vis}.
Additionally, we cut off the first and last 30 seconds of each video to remove any solicitations or other edited footage.

\noindent
\textbf{Data format.}
We segmented the video data into images at a rate of 10Hz, with a resolution of 1080p (1960$\times$1080) for each image.

\noindent
\textbf{Data cleaning.}
To automate the process of filtering out low-quality images from the dataset, we utilize a vision language model (VLM), LLaMA-Adapter V2~\cite{gao2023llamaadapterv2}.
First, we group the images at a rate of 2Hz and randomly select one image from each group to feed into the VLM. We provide a checklist, asking the VLM sequenced questions about the image quality. The checklist includes items such as non-front view, video transition, black screens, \textit{etc}. The VLM then uses its acquired world knowledge to infer and assist in automatically eliminating low-quality images. The checklist is organized as a set of text prompts given to the VLM, specified below.
{
\captionsetup{type=table}
\begin{tcolorbox}[colback=gray!10,
                  colframe=black,
                  width=\textwidth,
                  arc=1mm, auto outer arc,
                  boxrule=0.5pt,
                 ]
\texttt{\textcolor{blue}{Text Prompt Examples:}}

{\color{gray}
\texttt{"}Is the driving scenario image presented from a POV or other perspective rather than a front view?\texttt{"},

\texttt{"}Does the driving scenario image exhibit a gradual transition due to a video transition?\texttt{"},

\texttt{"}Is the image almost completely black, distinguishable from those depicting night driving?\texttt{"}, 

\texttt{"}Is this image excessively blurry, rendering any foreground object information indistinguishable?\texttt{"}, 

\texttt{"}Does this image feature subtitles, distinct from signs or markers in driving scenarios?\texttt{"}, 

\texttt{"}Does this image depict a scenic view or a bird's-eye perspective, distinct from front-view driving footage?\texttt{"}, \textit{etc.}
}
\end{tcolorbox}
}

\subsubsection{Multimodal Annotation}

Our OpenDV-Wild features three types of annotations: text, depth, and semantics.
We leverage the established BLIP2-flant5~\cite{li2023blip2} to describe each frame's main objects or scenarios with the following prompt. 
{
\captionsetup{type=table}
\begin{tcolorbox}[colback=gray!10,
                  colframe=black,
                  width=\textwidth,
                  arc=1mm, auto outer arc,
                  boxrule=0.5pt,
                 ]
\texttt{\textcolor{blue}{Text Prompt Example:}}

\texttt{"}Question: Describe the image of a driving scenario concisely. Answer:\texttt{"}.

\end{tcolorbox}
}    
We present a content query to the VLM according to each example's text prompts. 
If the VLM responds with \texttt{no} to all the queries, then the set of images can successfully pass the review.

For depth and semantic segmentation, we employ pre-trained ZoEDepth~\cite{bhat2023zoedepth} and Segformer~\cite{xie2021segformer} for automated label generation.
Segformer is previously trained on the CityScapes dataset~\cite{cordts2016cityscapes}.
The examples of text annotations are shown as follows.
{
\captionsetup{type=table}
\begin{tcolorbox}[colback=gray!10,
                  colframe=black,
                  width=\textwidth,
                  arc=1mm, auto outer arc,
                  boxrule=0.5pt,
                 ]
\texttt{\textcolor{blue}{Text Annotation Examples:}}

{\color{gray}
\texttt{"}A car is driving down a street in Rio De Janeiro.\texttt{"},

\texttt{"}A motorcyclist on a road with a speed limit sign.\texttt{"},

\texttt{"}A car driving down a street in Madrid, Spain.\texttt{"}, 

\texttt{"}A view of a city street with tall buildings and a TV tower in the background.\texttt{"}, 

\texttt{"}A snowy road with trees on both sides of the road and a red car is driving.\texttt{"}, 

\texttt{"}A car driving on a highway at dusk in Las Vegas, Nevada.\texttt{"}, \textit{etc.}
}
\end{tcolorbox}
}

\begin{table}[t]
\caption{\textbf{Location distribution of nuScenes and DIVA-Real}.}
\centering
\begin{tabular}{lccccc}
\toprule
Dataset & North America  & South America & Europe & Asian & Africa  \\
\midrule
nuScenes & 44.1\% & 0.0\% & 0.0\% & 55.9\% & 0.0\% \\
DIVA-Real & 56.9\% & 8.5\% & 16.9\% & 14.6\% & 3.1\% \\
\bottomrule
\end{tabular}
\label{tab:location_distribution}
\end{table}

\begin{table}[t]
\caption{\textbf{Time period distribution of nuScenes and DIVA-Real.}}
\centering
\begin{tabular}{lcccc}
\toprule
Dataset & Daytime & Dawn & Dusk & Nighttime  \\
\midrule
nuScenes & 88.4\% & 0.0\% & 0.0\% & 11.6\% \\
DIVA-Real & 55.8\% & 16.3\% & 10.1\% & 17.8\% \\
\bottomrule
\end{tabular}
\label{tab:period_distribution}
\end{table}

\begin{table}[t]
\caption{\textbf{Weather distribution of nuScenes and DIVA-Real}.}
\centering
\begin{tabular}{lccccc}
\toprule
Dataset & Normal & Rainy & Cloudy & Foggy & Snowy  \\
\midrule
nuScenes & 80.5\% & 19.5\% & 0.0\% & 0.0\% & 0.0\% \\
DIVA-Real & 58.2\% & 1.0\% & 28.6\% & 2.1\% & 10.2\% \\
\bottomrule
\end{tabular}
\label{tab:weather_distribution}
\end{table}

\subsubsection{Appearance Diversity Highlights}
\label{sec:diversity}

\noindent 
\textbf{Diversity over prior datasets.}
Beyond its large data scale, our dataset outshines competitors in terms of \textit{appearance diversity}.
YouTube, known for its diverse content, is regarded as a global mosaic and a crucial source of data.
DIVA-Real leverages this by comprising 120 hours of publicly available videos from over 71 cities across more than 19 countries.
Our dataset exhibits a globe-wise geographic distribution compared to other open datasets collected in limited regions.
On top of that, DIVA-Real covers a rich variety of driving scenarios, including bridges, bays, wilderness, deserts, dusk, fog, and more.
Unlike datasets restricted to dull scenes, our dataset empowers models to capture a wealth of visual appearance diversity.
Lastly, ours boasts a richness of annotations on par with other datasets.

In this section, we provide a detailed data analysis about the diversity of DIVA-Real. 
For simplicity's sake, we assume that all clips within a video are shot at the same location and time, with any single frame from a segment representative of the geographical location, lighting conditions, weather, and other informational aspects concerning the video.
Thus, we manually review each video's title and a random frame from the video and assess its geographic location, time, and weather conditions, among other things, for statistical analysis. The following diversity analysis has been derived from this process in three aspects.

\noindent
\textbf{Location distribution.}
According to statistical results, YouTube videos are derived from 71 cities in 19 countries, covering a much larger area than any existing public driving dataset, as shown in \cref{tab:location_distribution}. For example, in the most popular regions, DIVA-Real includes 67 hours of video data in the United States, covering cities like Los Angeles, New York, Las Vegas, Miami, Boston, Atlanta, New Orleans, \textit{etc.}, and encompasses geographical areas such as urban, rural, coastal, wilderness, mountainous, and port regions.

\noindent
\textbf{Time period and weather variance.}
The DIVA-Real dataset also includes a variety of times and weather conditions. As shown in \cref{tab:period_distribution}, in addition to daytime, the dataset covers a considerable proportion of dawn, dusk, and nighttime scenarios. \cref{tab:weather_distribution} presents the weather distribution in the dataset, including rainy, cloudy, foggy, and snowy conditions.
These diverse times and weather conditions ensure a variety of appearances.

\noindent
\textbf{Corner cases.}
YouTube videos also include extreme cases and safety-critical scenarios. \cref{fig:diva_vis} presents several special cases from DIVA-Real, such as crowded pedestrian-filled intersections (fourth one in the first row), roads with a lot of parked vehicles (fifth one in the first row), a country road in the sunset (third one in the fifth row), and passing under an overhead structure with limited light (fifth one in the seventh row).

\subsection{DIVA-Sim}

\subsubsection{Data Collection}

The DIVA-Sim dataset is collected through the MetaDrive simulator~\cite{li2022metadrive}. 
%
DIVA-Sim accumulated a total of 27.5 hours of virtual driving data, including 5.5 hours from the digital twins of the nuScenes dataset~\cite{caesar2019nuscenes} via ScenarioNet~\cite{li2023scenarionet}, and 22 hours of dangerous driving scenarios collected initially based on the Waymo Open dataset~\cite{sun2020scalability} by adversarial interventions~\cite{zhang2023cat}.
For data from nuScenes and Waymo Open, each video lasts 20 seconds and 8 seconds, respectively.

For each scenario, DIVA-Sim provides a variety of labels, including rendered RGB, depth, segmentation, instance map, and top-down view.
All camera outputs are synthesized using the OpenGL rendering backend from the Panda3D game engine, allowing us to incorporate depth map and semantic colormap similar to ZoeDepth~\cite{bhat2023zoedepth} and Segformer~\cite{xie2021segformer}.

\noindent
\textbf{Data format.}
We segmented the video data into images at a rate of 10Hz, with a resolution of 1960$\times$1080 for each image.

\subsubsection{Layout Diversity Highlights}
\begin{figure}
    \centering
    \includegraphics[width=\linewidth]{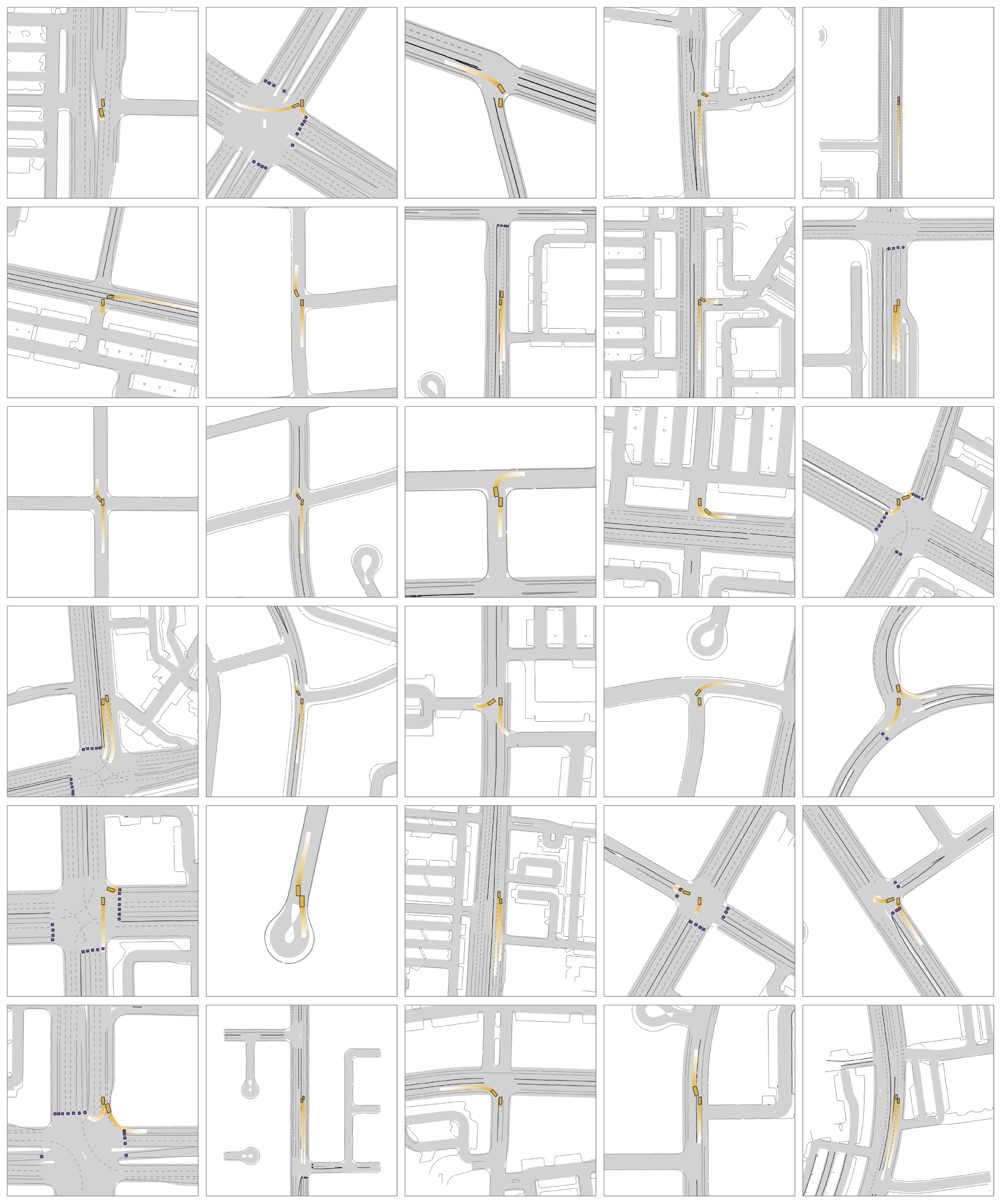}
    \caption{\textbf{Various safety-critical layouts from DIVA-Sim.}
    The yellow rectangular represents the ego-car and the other cars interacting with it.
    All scenarios are initialized by the Waymo Open dataset~\cite{sun2020scalability} and generated by adversarial interactions~\cite{zhang2023cat} within simulators.
    }
    \label{fig:diva-sim}
\end{figure}

We randomly sample 500 safety-critical videos generated based on Waymo Open dataset~\cite{sun2020scalability} and manually reviewed the contents of the top-down view of each video, collecting the data shown in \cref{tab:layout_distribution}.
Beyond forwarding, turning, and stopping, DIVA-Sim also covers cases like changing lanes, passing through intersections, and making U-turns. 
\cref{fig:diva-sim} visually displays the top-down views of various dangerous driving scenarios, including collisions, quick stops, and reckless merging. 
These illustrate the diversity of layouts in DIVA-Sim.

\begin{table}[t]
\caption{\textbf{Layout distribution of nuScenes and DIVA-Sim}.}
\centering
\tablestyle{2.0pt}{1.05}
\setlength{\tabcolsep}{1mm}{
\begin{tabular}{lcccccccc}
\toprule
\multirow{2}{*}{Dataset}  & \multirow{2}{*}{Forward} & Left & Right & Left Lane & Right Lane & Intersection & \multirow{2}{*}{U-Turn} & \multirow{2}{*}{Stop}\\
 & & Turn & Turn & Change & Change & Passing & & \\
\midrule
nuScenes & 47.1\% & 18.0\% & 10.2\% & 5.0\% & 2.5\% & 13.1\%  & 0.0\% & 4.1\% \\
DIVA-Sim & 36.2\% & 14.3\% & 10.0\% & 10.7\% & 17.4\% & 6.6\%  & 1.2\% & 3.6\% \\
\bottomrule
\end{tabular}}
\label{tab:layout_distribution}
\end{table}

\subsection{Extensions on Public Dataset}

In addition to collecting YouTube data and simulation data, we also annotate data of public datasets to promote research from simulation to reality and to enable fair comparisons.

\noindent
\textbf{nuScenes dataset.}
The nuScenes dataset~\cite{caesar2019nuscenes} is a public driving dataset that includes 1000 scenes from Boston and Singapore for diverse driving tasks~\cite{zhou2021tempnet,liu2023density,liu2023apr}. Each scene comprises a 20-second video, approximately 40 frames. It provides 700 training scenes, 150 validation scenes, and 150 test scenes. Similarly, we utilize BLIP2-flant5, ZoeDepth, and Segformer to provide textual, depth, and semantic labels for this dataset.

\subsection{License and Privacy Considerations}
\label{sec:license}


\setcounter{footnote}{0}
All the data is under the CC BY-NC-SA 4.0 license\footnote{https://creativecommons.org/licenses/by-nc-sa/4.0/deed.en}.
Other datasets (including nuScenes \cite{qian2023nuscenes}, Waymo Open \cite{sun2020scalability}, Metadrive~\cite{li2022metadrive}) inherit their own distribution licenses.

We place a high value on license and privacy protection, following the precedent from YouTube-8M \cite{abu2016youtube}, YouTube-VOS \cite{xu2018youtube}, AOC \cite{zhang2022learning}, CelebV-HQ \cite{zhu2022celebvhq}, ELM~\cite{zhou2024embodied}, OpenScene~\cite{contributors2023openscene}, and Kinetics \cite{kay2017kinetics}, \textit{etc}. 
For videos from YouTube, permission to access the video content is received through a Creative Commons license.
Besides, we skip channel-related content at the beginning and end of the videos during data processing to ensure we do not infringe upon the rights of logos, channel owner information, or other copyrighted materials.
We do not provide video content; users are redirected to original YouTube videos via a link. 
The platform safeguards personal info with encryption, access limits, and identity checks to prevent unauthorized video access. 
We will credit the source, provide a link to the license, and state that no modifications have been made to the video itself, and the data will not be used for commercial purposes.
All the data we obtain complies with regulations and YouTube's Privacy Policy.
In addition, we comply with any limitations required by applicable law and any requests submitted by users. For instance, users may have the right to view, correct, and delete personal information we possess about them, such as deleting text labels, unlinking videos, and de-identifying data.
%

\section{Implementation Details of SimGen}
\label{sec:implementation}

\subsection{Empirical Study}
\label{sec:empirical study}

\subsubsection{Cascade Diffusion Scheme}
\label{sec:cascade}

\begin{figure}[t]
    \centering
    \includegraphics[width=0.85\linewidth]{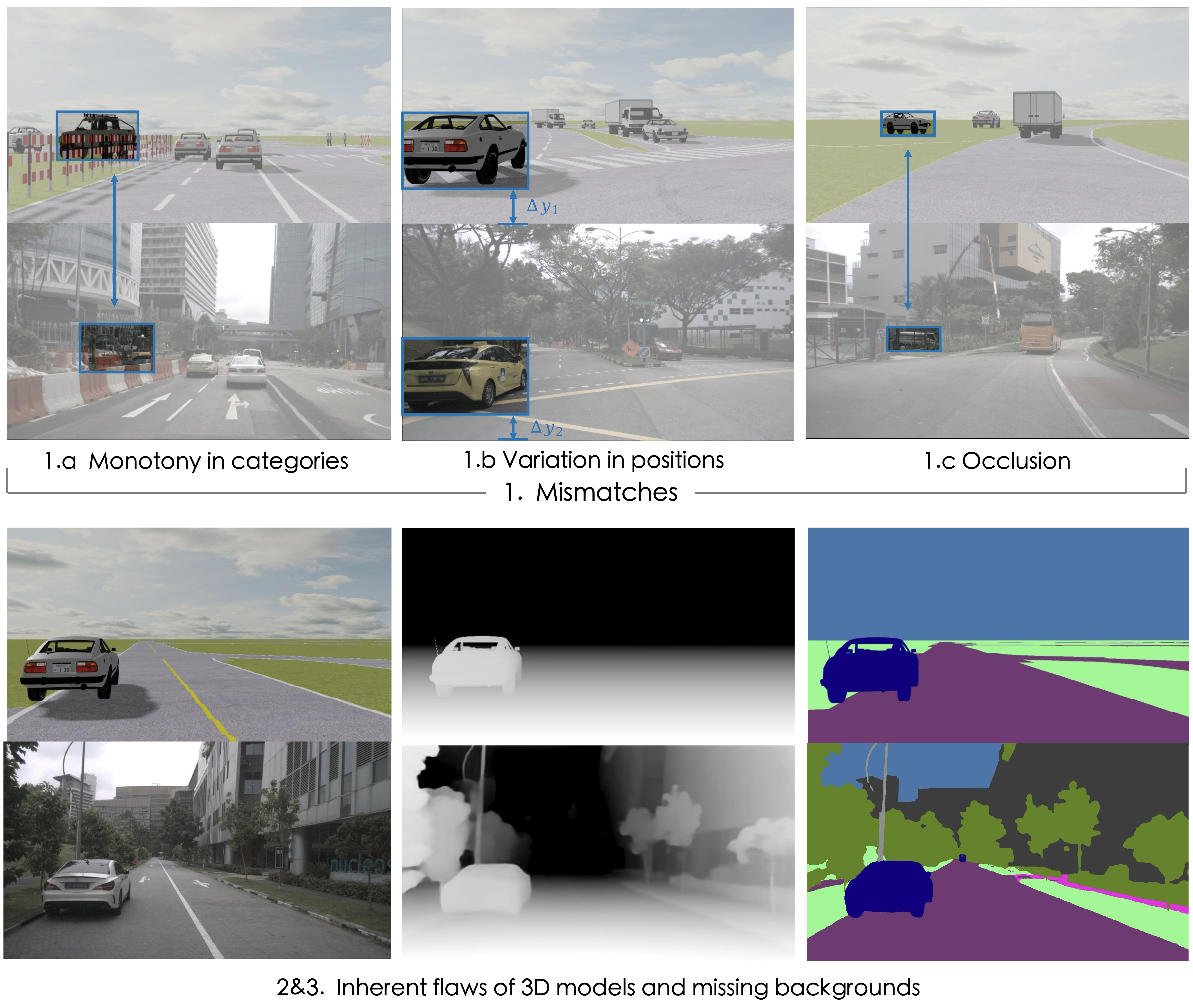}
    \caption{\textbf{Gaps between conditions in simulators and reality.}
    In each group of images, the first row represents the conditions in the simulator, and the second row represents the real conditions.
    }
    \label{fig:gaps}
\end{figure}

It's worth noting that, despite our efforts to replicate real scenes as closely as possible in the simulator, gaps are inevitably present.
Gaps between the conditions in simulators and reality can affect the accuracy of model training. 
These gaps primarily originate from: 1) mismatches, 2) inherent flaws of 3D models, and 3) missing backgrounds.
In \cref{fig:gaps}, we provide some visualizations to illustrate these gaps.

The mismatch between simulated and real images can exist in multiple aspects.
Firstly, given that the models within the simulator are finite, it is unrealistic to accurately represent a wide variety of object categories through it. As shown in example 1.a, for excavators, the simulator uses dump trucks based on size, resulting in some mismatches in object shape. 
Secondly, slight differences between the camera positions in the simulator and real datasets can cause objects to be displaced in the simulator, as in example 1.b, where $\Delta y_1$ is larger than $\Delta y_2$. 
Thirdly, the simulator lacks physical entities such as buildings or trees, making it difficult to replicate scenarios where vehicles are obstructed by the environment, as shown in example 1.c.
Beyond mismatches, inherent flaws may also exist within the simulator models. Although the simulator can reflect the real size of vehicles, achieving perfect consistency in shape is still challenging. Furthermore, the transparency of the model's window parts can lead to discrepancies between depth and semantic segmentation results in those areas.
Finally, the simulator lacks elements such as buildings, trees, and street lamps, which are essential conditions for controlling image generation.

Through the above analysis and visualization results, combined with the outcomes of our ablation study, we infer that real depth and semantic segmentation images are more suited to control real image generation than simulated equivalents. Therefore, SimGen's cascade diffusion network is necessary, as it allows the decoupling of the introduction of SimCond in the CondDiff module from the controlled image generation in the ImgDiff module.

It's worth noting that this gap doesn't mean simulated conditions can't be used for training image generation models. Instead, it suggests that employing real depth and semantics is a better choice, and better aligned for joint training with data from YouTube. Indeed, the simulator's ExtraCond can also contribute to the model's accuracy. For instance, the instance map can indicate the object count and occlusion relationships in the scene to the model, and the top-down view can provide spatial location information. These insights can guide the use of a unified adapter to merge multi-modal conditions.

\vspace{-5pt}
\subsubsection{Unified Adapter}
\label{sec:adapter}
\vspace{-5pt}

SimGen employs a unified adapter to address the two obstacles in multi-modal condition conflicts: \textit{Modal Discrepancy} and \textit{Condition Disparity}. Modal discrepancy refers to the inconsistent number of modalities between data from nuScenes and YouTube (the latter lacks ExtraCond), which might lead the model to establish a statistical shortcut, such as outputting nuScenes-style images when ExtraCond is present and YouTube-style images when it's absent. 
This shortcut can significantly impact the model's instruct-following ability and diversity at inference time. 
On the other hand, condition disparity refers to the lack of background information in the ExtraCond condition, which can result in conflicting control information with RealCond. 
Thus, our proposed solution is to use an adapter to merge various modalities into a unified control feature, employing a mask during the fusion process to eliminate conflicts arising from absent background information in ExtraCond.

\vspace{-5pt}
\subsection{Model Design}
\vspace{-5pt}



\smallskip
\noindent
\textbf{Realism-controllability trade-off.}
Apart from the discretization steps of the CondDiff solver, the critical hyperparameter for sim-to-real transformation is $t_s$, the starting time of the image synthesis process in the reverse SDE.
We notice that with a fixed CondDiff model, there's a trade-off between Realism and Controllability when choosing different $t_s$ values. 
Smaller $t_s$ values lead to fewer denoising steps, giving SimCond more control over image generation but potentially compromising realism. 
Generally, we find $t_s\in [0.4,0.65]$ to work well for the foreground, and we ultimately select $t_s$ as 0.5 for foregrounds.

\textbf{Extension on Video Generation}
Having acquired the single-frame variant, we lock the original blocks within the denoising UNet and intersperse them with temporal reasoning blocks, mirroring the strategy of GenAD~\cite{yang2024generalized}, thereby facilitating video sequence modeling.

\subsection{Training Details}

SimGen is trained in two modules: CondDiff, which converts simulated conditions to real ones, and ImgDiff, which generates images from multimodal conditions.
In the first stage, we fine-tune the pre-trained SD-2.1-V on per-image denoising with 1.1B trainable parameters of its denoising UNet. 
It is trained on 4.5M text-depth-segmentation pairs of DIVA-Real and nuScenes. 
We train the model for 30K iterations on 8 GPUs with a batch size of 96 with AdamW~\cite{loshchilov2018fixing}.
We linearly warm up the learning rate for $10^3$ steps in the beginning, then keep it constant at $1 \times 10^{-5}$. 
The default GPUs in most of our experiments are NVIDIA Tesla A6000 devices unless otherwise specified.

In the second stage, we train the model via a unified adapter and ControlNet using text-condition-image pairs, lifting it to generate realistic images during inference.
The training data consists of DIVA and nuScenes, with conditions confined to ExtraCond and RealCond.
Following the design of ControlNet, we freeze the input and middle layers of the UNet, training only the parameters of the control branch and the Adapter.
This stage is trained for 50,000 iterations on 8 GPUs, with a 295 batch size of 96.

For the extension of video generation, we freeze all blocks of the single-frame version and only optimize the introduced temporal reasoning blocks, resulting in 418M trainable parameters in this stage.
To maximize the data efficiency for constructing video clips, we take each frame of a 10Hz YouTube and nuScenes video as a starting frame to form a 3s training sequence at 2Hz.
The text condition is structured in the same way as the first stage, and we acquire the context from the middle frame of the sequence.
SimGen is trained on 8 GPUs for 30K iterations with a total batch size of 24. 
The learning rate is set as $1 \times 10^{-5}$ after $10^3$ warm-up steps.

In both stages, the input frames are resized to $256\times448$, and the text condition $c$ is dropped at a probability of $\gamma_c = 0.1$ to enable classifier-free guidance~\cite{Ho2021classifierfree} in sampling. 
Both CLIP text encoders and the autoencoder are kept frozen throughout our experiments.
%


%
For effective classifier-free guidance~\cite{Ho2021classifierfree}, ImgDiff random drops conditions during training at a rate of $\gamma_c=0.1$.
Additionally, we enhance the model's robustness by randomly masking the background of real conditions with a fixed probability of $\gamma_b=0.5$.
To address potential cumulative errors from the CondDiff process during ImgDiff denosing, we introduce slice noise with a probability of $\gamma_n=0.25$. 
Slice noise entails partitioning the image into $n\times n$ patches and randomly masking them with a probability of $\gamma_p=0.25$.

\vspace{-5pt}
\subsection{Sampling Details}
\vspace{-5pt}

Given conditions from simulators, SimGen first has a reverse SDE process in CondDiff.
It starts from SimCond added with standard Gaussian noises.
The sampling step of this stage is 25 ($t_s=0.5$).
After that, the second sampling process is involved in ImgDiff, starting with random Gaussian noises.
Both sampling processes are performed by Denoising Diffusion Implicit Models (DDIM)~\cite{song2020DDIM}.  
We use 50 sampling steps and set the scale of classifier-free guidance to 9.5. 
The sampling speed is 1.13 seconds per step per batch.
The image resolution is $256\times448$, and the video sequence is at 2Hz.

\vspace{-5pt}
\section{Experiments}
\label{sec:supp_exp}
\vspace{-5pt}


We conduct extensive experiments on multiple datasets to evaluate the performance of our method.
For comparison convenience, we trained two models on the nuScenes and DIVA datasets, respectively, namely SimGen-nuSc and SimGen, adopting the same training strategy.

\subsection{Metrics}

One of the roles of synthesized images is to augment existing perception models of autonomous driving~\cite{zhou2021monocular,zhou2021monoef,zhou2023monoatt,zhou2022mogde}.
We use various metrics in multiple aspects for quantitative evaluation. 
For generation quality metrics, we use Fréchet Inception Distance (FID)~\cite{Heusel2017FID} and Fréchet Video Distance (FVD)~\cite{Unterthiner2018FVD}.
For generation diversity, the pixel diversity $D_{\text{pix}}$ metric is leveraged.
The controllability of the model is reflected by the alignment between the generated image and the conditioned BEV sequences.
For the video generation task, all frames are at 2Hz.
The specific metrics are described as follows.

\smallskip
\noindent\textbf{FID:} 
It evaluates the generation quality of images, which are video frames in our experiments, by measuring the distribution distance of features between the predictions and original frames in the dataset. The features are extracted by a pre-trained Inception model. For quantitative comparison on nuScenes, FID is evaluated on 6019 generated frames and ground-truth frames. For experiments on YouTube, FID is calculated on 18000 frames from both generation and the dataset.

\smallskip
\noindent\textbf{FVD:} It measures the semantic similarity between real and synthesized videos with a pre-trained I3D action classification model~\cite{Carreira2017I3D} as the feature extractor. We evaluate 4369 video clips from nuScenes and 3000 video clips from YouTube.

\smallskip
\noindent\textbf{$D_{\text{pix}}$:} To gauge the diversity of the generated data, we compute the standard deviation of the pixel values in the generated images. A higher value indicates a greater diversity of colors in the generated data.
The conditions for the evaluation come from the nuScenes validation set, and text prompts are collected from both the nuScnes validation set and randomly selected 18,000 frames of YouTube data. 
For each condition, we randomly select a text from the collected prompts as input, and the model generates the corresponding image. 
To reduce randomness, we test each model 3 times with the same random seed and take the average.

\begin{figure}[t]
    \centering
    \includegraphics[width=0.85\linewidth]{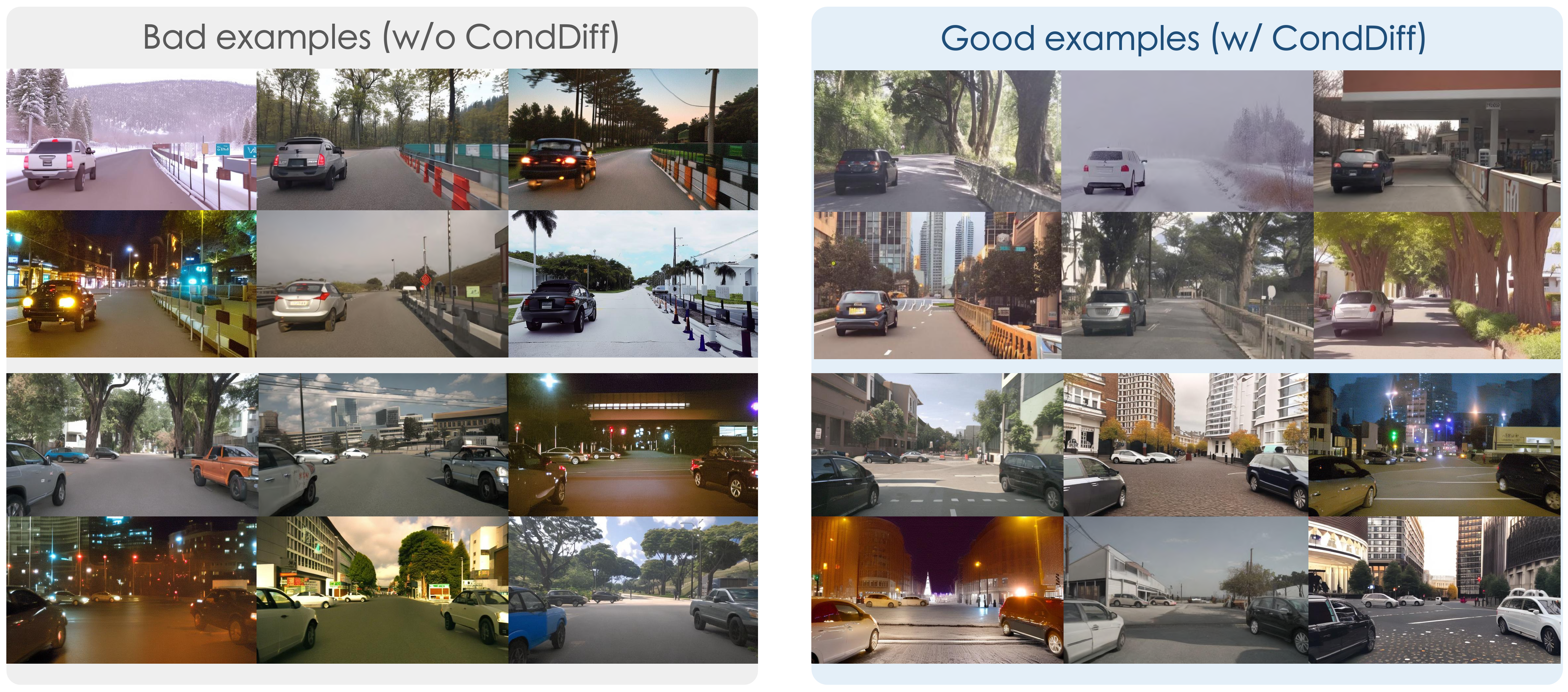}
    \caption{\textbf{Comparison of proposed cascade diffusion model (blue boxes) to na\"ive approaches (gray boxes).}}
    \label{fig:supp_cond}
    \vspace{-7pt}
\end{figure}

\begin{figure}[t]
    \centering
    \includegraphics[width=\linewidth]{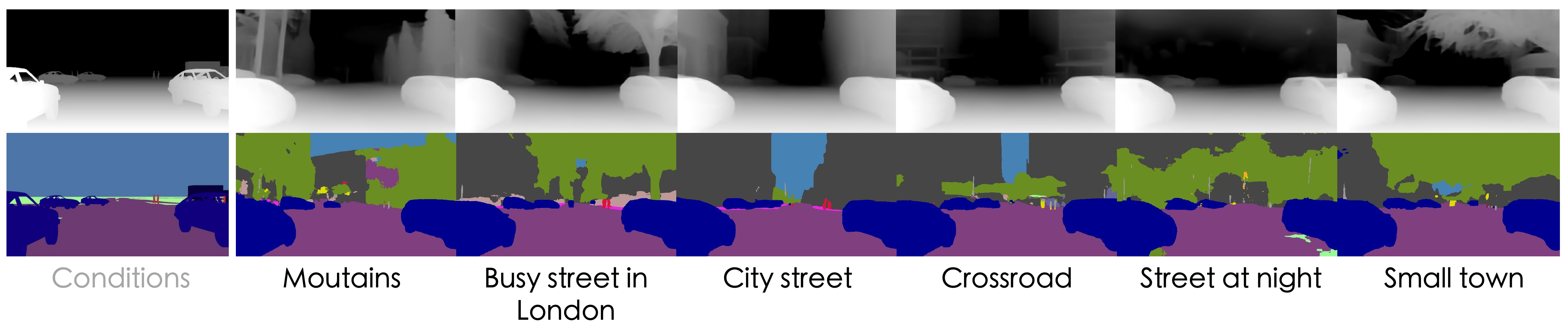}
    \caption{\textbf{Visualization of text-grounded Sim-to-Real condition transformation.}}
    \label{fig:conddiff}
    \vspace{-7pt}
\end{figure}

\begin{figure}[t]
    \centering
    \includegraphics[width=\linewidth]{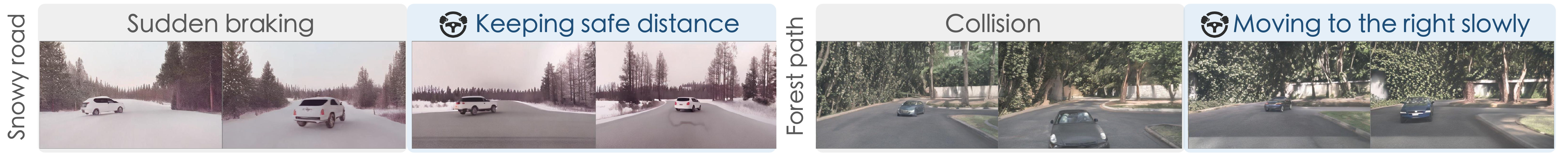}
    \caption{\textbf{Preliminary attempt at closed-loop evaluation.}
    For two scenarios, the IDM behavior~\cite{li2022metadrive} (gray boxes) leads to hazardous driving situations, while manual control (blue boxes) adopts measures to evade risks.
    }
    \label{fig:closed-loop}
    \vspace{-7pt}
\end{figure}

\smallskip
\noindent\textbf{Controllability Metrics:}
To affirm the consistency between the generated images and original data in layout, we employ metrics such as Average Precision (AP), and Mean Intersection over Union (mIoU) to evaluate the perception performance on the nuScenes dataset. Our evaluation comprises two aspects: firstly, we use pre-trained perception models to compare the validating performance of generated data with real data. Secondly, we explore the potential of employing augmented training sets as a strategy for performance improvement. 

We adopt BEVFusion~\cite{liang2022bevfusion}, a state-of-the-art perception method, as our primary evaluation tool. Specifically, we utilize a BEVFusion implementation that only incorporates a front view, masking other camera perspectives and ground truth during evaluation.

\vspace{-5pt}
\subsection{More Ablations}
\label{sec:more_ablation}
\vspace{-5pt}

\begin{wrapfigure}{r}{0.4\textwidth}
\begin{minipage}[b]{0.4\textwidth}
\vspace{-12pt}
\captionof{table}{\textbf{Quality of video generation.}} 
\footnotesize
\centering
\tablestyle{2.0pt}{1.05}
\setlength{\tabcolsep}{1mm}{
\begin{tabular}{lc}
\toprule
Method   & FVD$\downarrow$  \\ \midrule
DriveGAN~\cite{kim2021drivegan}   & 502             \\
DriveDreamer~\cite{wang2023drivedreamer} & 452  \\
DrivingDiffusion~\cite{li2023drivingdiffusion} &  332  \\
GenAD~\cite{yang2024generalized} & \textbf{184} \\ \midrule
SimGen  & \textit{271}            \\ \bottomrule
\end{tabular}
}
\label{tab:fvd}
\vspace{-5pt}
\end{minipage}
\end{wrapfigure}

\textbf{Results of video generation.}
It's worth noting that the innovation of SimGen doesn't focus on video generation. 
However, high-quality image generation implies the potential for video generation, which is crucial for interactive scenario generation and closed-loop planning. 
We made a preliminary attempt at video generation based on GenAD~\cite{yang2024generalized}. \cref{tab:fvd} compares SimGen with other video generation models. Thanks to its commendable image generation quality, SimGen achieves performance that is on par with other models.

\textbf{Effectiveness of Sim2Real condition transformation.}
To validate the effectiveness of the cascade diffusion model, we display a set of comparative images in \cref{fig:supp_cond}.
The blue boxes represent the cascade diffusion structure we adopt, it transforms SimCond from the simulator into RealCond in CondDiff, and then generates realistic images through the ImgDiff model. The grey boxes signify that we removed CondDiff, and ImgDiff directly generates images using SimCond.
It is observable that removing CondDiff causes the generative model to introduce certain distortions from the simulator conditions into the produced images, as seen in the overly narrow wheels and deformed rear part shown in the left set of pictures. 
The introduction of CondDiff transforms these distortions into realistic conditions after resampling, thus greatly enhancing image generation quality.
\cref{fig:conddiff} further demonstrates that CondDiff can transform depth and segmentation from the simulator into real ones based on different texts.

\textbf{Closed-loop evaluation.}
%
We further explore applying our simulator-conditioned generative models to the closed-loop evaluation in \cref{fig:closed-loop}.
The evaluation focuses on two driving behaviors, namely IDM~\cite{li2022metadrive} (gray boxes) and manual control (blue boxes) in different scenarios.
IDM could lead to risks like sudden braking or collision in these cases.
Conversely, manual control promotes safety by maintaining distance and slowing down.
The video data is generated by SimGen, using conditions pulled from simulator interactions, with a one-second frame interval.

\begin{figure}[H]
    \centering
    \includegraphics[width=0.95\linewidth]{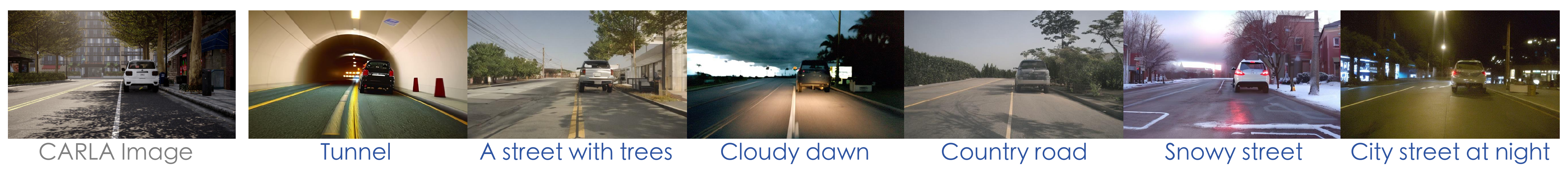}
    \caption{\textbf{Case study of zero-shot image generation on CARLA.}
    We randomly select a scenario in CARLA, for which SimGen generates various driving scenarios through the conditions (depth and segmentation) produced by the simulator and different textual prompts.
    }
    \label{fig:carla}
\end{figure}

\textbf{Generalization on novel simulators.}
As CondDiff can convert simulated conditions into real conditions in an adaptation-free approach, SimGen possesses the ability to perform zero-shot generalization to other simulators. 
In \cref{fig:carla}, we exhibit a case study of generating realistic images using depth and semantic segmentation conditions provided by CARLA~\cite{dosovitskiy2017carla}.
This provides the possibility for SimGen to utilize and integrate the diverse layouts, driving policies, and physical engines provided by various simulation platforms to generate diverse driving scenarios.

\vspace{-5pt}
\subsection{Qualitative Results}
\vspace{-5pt}

\begin{figure}[H]
    \centering
    \includegraphics[width=0.95\linewidth]{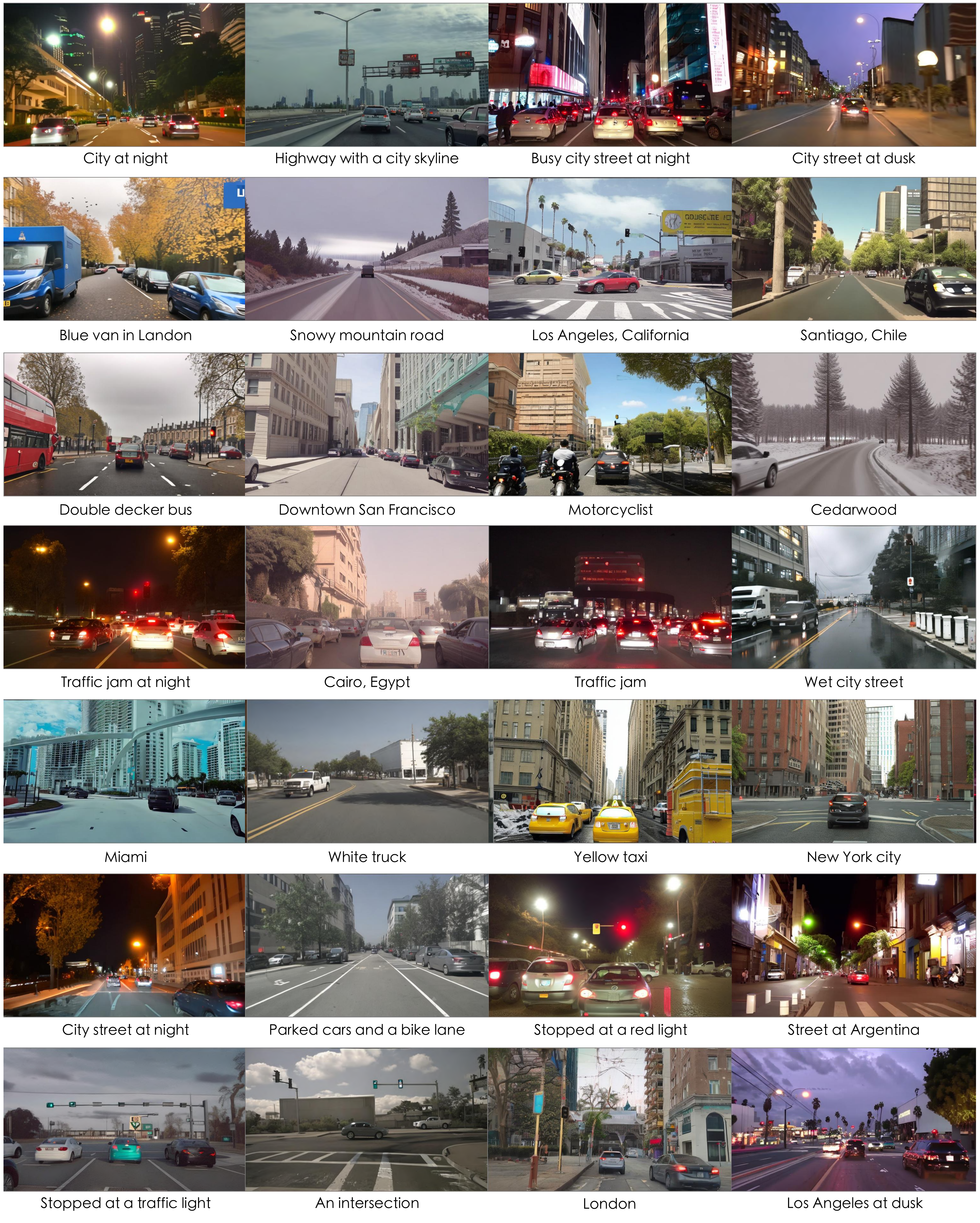}
    \caption{\textbf{Text-grounded image generation.}
    Each image is generated by SimGen using a randomly selected text prompt and simulator conditions.
    The rich appearance diversity is reflected through the wide range of generated content.
    }
    \label{fig:text-prompt}
\end{figure}

\textbf{Text-grounded image generation.}
SimGen is a capable text-to-image diffusion model for driving scenarios, especially when examining text controllability compared to other works. In \cref{fig:text-prompt}, we demonstrate SimGen's exceptional ability to generate images from different text prompts. Thanks to the relatively simple simulator conditions and comprehensive \textbf{DIVA} dataset, the text prompt can effectively influence the resulting image, even changing the surrounding building and background with reference to specific cities. Whereas as other diffusion-based generative models struggle to generate images with characteristics that were not originally present in nuScenes ~\cite{gao2023magicdrive}, like unseen weather or background settings, Simgen's text-grounding can influence both foreground objects like cars and match the background to cities not present in nuScenes.

\begin{figure}[H]
    \centering
    \includegraphics[width=\linewidth]{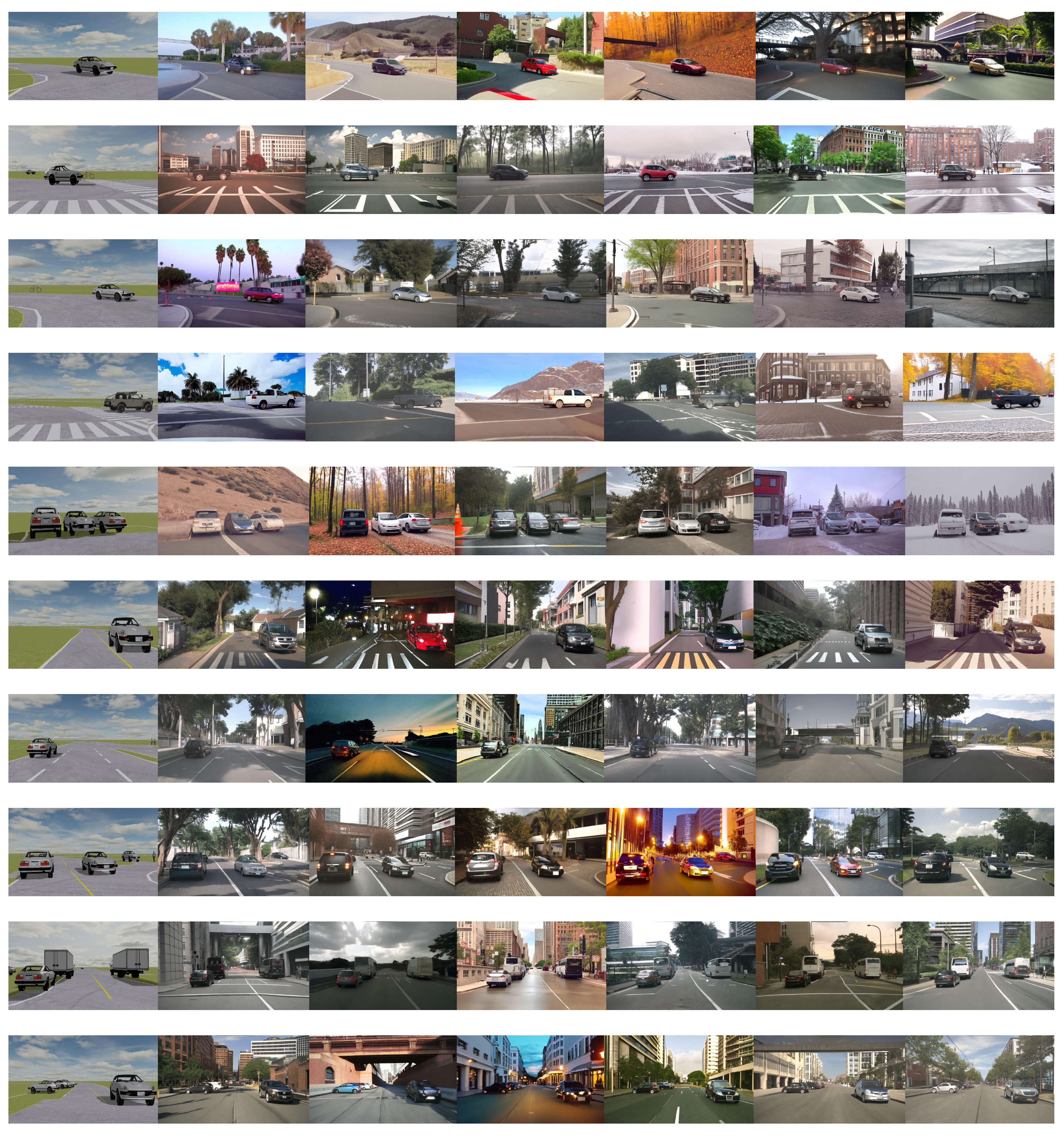}
    \caption{\textbf{Diverse generated images from simulator scenarios} From the same original simulated driving scenario (left column), we show a diverse range of generated images (columns 2 through 7). SimGen is capable of generating driving scenes in a wide variety of settings based on the same simulator conditions.}
    \label{fig:sim-to-real-diverse}
\end{figure}

\textbf{Simulator-conditioned image generation.}
\cref{fig:sim-to-real-diverse} displays additional examples of SimGen generating diverse images based on conditions provided by the simulator. The far-left column presents the simulator-rendered RGB, while the right side shows images generated by SimGen following different text prompts. This further validates SimGen's potent ability to adhere to the simulator conditions while maintaining rich appearance diversity.

\begin{figure}[H]
    \centering
    \includegraphics[width=\linewidth]{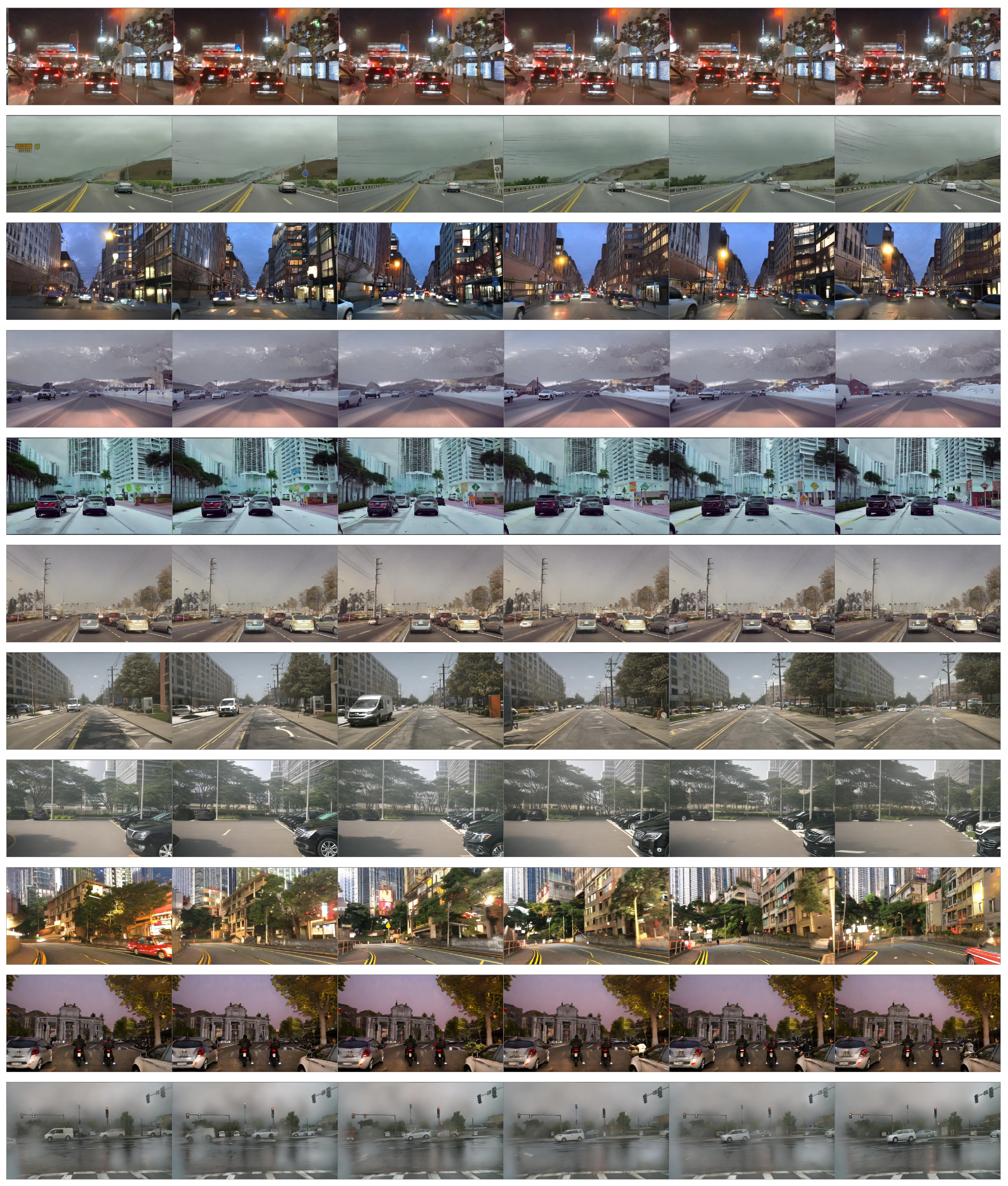}
    \caption{\textbf{Preliminary attempt at video generation.}
    Notably, SimGen is not designed for video generation.
    We simply follow some practices in \cite{yang2024generalized} to temporal consistency, and video generation will be our future work.
    }
    \label{fig:video}
\end{figure}

\textbf{Video generation.}
Our preliminary SimGen video generation model is able to maintain the ability to create driving images with a wide range of backgrounds while also incorporating temporal consistency as shown in \cref{fig:video}.

\vspace{-5pt}
\subsection{Failure Cases}
\vspace{-5pt}

\begin{figure}[H]
    \centering
    \includegraphics[width=0.98\linewidth]{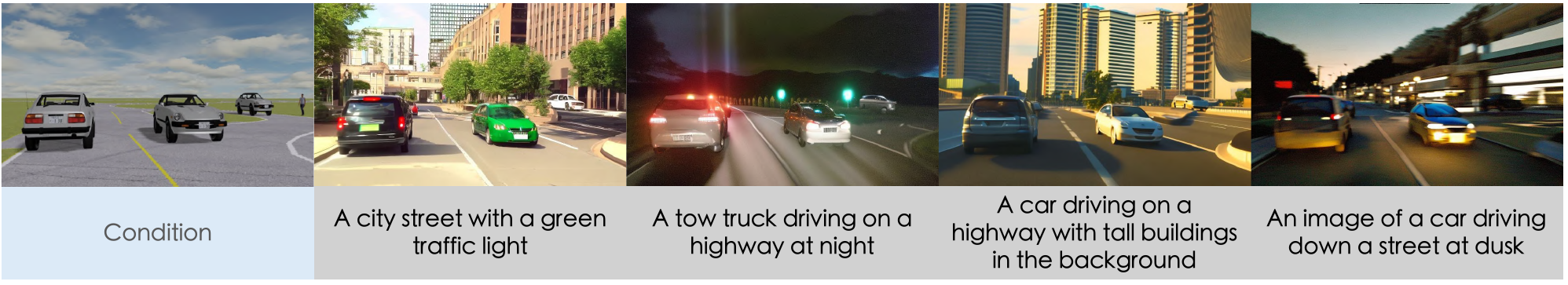}
    \caption{\textbf{Failure cases of SimGen.}}
    \label{fig:failure-cases}
\end{figure}

We show some failure cases of SimGen in \cref{fig:failure-cases}.
The first column represents the conditions from the simulator, with each generated image accompanied by a text prompt.
%
The failure cases of SimGen are included as follows:
1) Text comprehension error: as in the first image, where the adjective "green" is not assigned to any discernible "traffic light" but instead as a vehicle color.
2) Condition conflict: the second image provides a text prompt of a tow truck, but it's challenging for the model to generate such a vehicle based on the shape of a sedan.
3) Background subsumption: the third image demonstrates a case where the background subsumes the smaller car. 
4) Generation instability: in the last image, SimGen occasionally produces distortion and blur in background generation, likely due to cumulative model error and overabundance of nighttime images in DIVA.
%


\end{document}


%% file: main.bbl
\begin{thebibliography}{91}
\providecommand{\natexlab}[1]{#1}
\providecommand{\url}[1]{\texttt{#1}}
\expandafter\ifx\csname urlstyle\endcsname\relax
  \providecommand{\doi}[1]{doi: #1}\else
  \providecommand{\doi}{doi: \begingroup \urlstyle{rm}\Url}\fi

\bibitem[Abu-El-Haija et~al.(2016)Abu-El-Haija, Kothari, Lee, Natsev, Toderici, Varadarajan, and Vijayanarasimhan]{abu2016youtube}
Sami Abu-El-Haija, Nisarg Kothari, Joonseok Lee, Paul Natsev, George Toderici, Balakrishnan Varadarajan, and Sudheendra Vijayanarasimhan.
\newblock Youtube-8m: A large-scale video classification benchmark.
\newblock \emph{arXiv preprint arXiv:1609.08675}, 2016.

\bibitem[Ajay et~al.(2024)Ajay, Han, Du, Li, Gupta, Jaakkola, Tenenbaum, Kaelbling, Srivastava, and Agrawal]{ajay2024compositional}
Anurag Ajay, Seungwook Han, Yilun Du, Shuang Li, Abhi Gupta, Tommi Jaakkola, Josh Tenenbaum, Leslie Kaelbling, Akash Srivastava, and Pulkit Agrawal.
\newblock Compositional foundation models for hierarchical planning.
\newblock \emph{Advances in Neural Information Processing Systems}, 36, 2024.

\bibitem[Bhat et~al.(2023)Bhat, Birkl, Wofk, Wonka, and M{\"u}ller]{bhat2023zoedepth}
Shariq~Farooq Bhat, Reiner Birkl, Diana Wofk, Peter Wonka, and Matthias M{\"u}ller.
\newblock Zoedepth: Zero-shot transfer by combining relative and metric depth.
\newblock \emph{arXiv preprint arXiv:2302.12288}, 2023.

\bibitem[Blattmann et~al.(2023{\natexlab{a}})Blattmann, Dockhorn, Kulal, Mendelevitch, Kilian, Lorenz, Levi, English, Voleti, Letts, et~al.]{blattmann2023stable}
Andreas Blattmann, Tim Dockhorn, Sumith Kulal, Daniel Mendelevitch, Maciej Kilian, Dominik Lorenz, Yam Levi, Zion English, Vikram Voleti, Adam Letts, et~al.
\newblock Stable video diffusion: Scaling latent video diffusion models to large datasets.
\newblock \emph{arXiv preprint arXiv:2311.15127}, 2023{\natexlab{a}}.

\bibitem[Blattmann et~al.(2023{\natexlab{b}})Blattmann, Rombach, Ling, Dockhorn, Kim, Fidler, and Kreis]{blattmann2023align}
Andreas Blattmann, Robin Rombach, Huan Ling, Tim Dockhorn, Seung~Wook Kim, Sanja Fidler, and Karsten Kreis.
\newblock Align your latents: High-resolution video synthesis with latent diffusion models.
\newblock In \emph{Proceedings of the IEEE/CVF Conference on Computer Vision and Pattern Recognition}, pages 22563--22575, 2023{\natexlab{b}}.

\bibitem[Caesar et~al.(2020)Caesar, Bankiti, Lang, Vora, Liong, Xu, Krishnan, Pan, Baldan, and Beijbom]{caesar2019nuscenes}
Holger Caesar, Varun Bankiti, Alex~H. Lang, Sourabh Vora, Venice~Erin Liong, Qiang Xu, Anush Krishnan, Yuxin Pan, Giancarlo Baldan, and Oscar Beijbom.
\newblock {nuScenes}: A multimodal dataset for autonomous driving.
\newblock 2020.

\bibitem[Caesar et~al.(2021)Caesar, Kabzan, Tan, Fong, Wolff, Lang, Fletcher, Beijbom, and Omari]{caesar2021nuplan}
Holger Caesar, Juraj Kabzan, Kok~Seang Tan, Whye~Kit Fong, Eric Wolff, Alex Lang, Luke Fletcher, Oscar Beijbom, and Sammy Omari.
\newblock nuplan: A closed-loop ml-based planning benchmark for autonomous vehicles.
\newblock \emph{arXiv preprint arXiv:2106.11810}, 2021.

\bibitem[Carreira and Zisserman(2017)]{Carreira2017I3D}
Joao Carreira and Andrew Zisserman.
\newblock Quo vadis, action recognition? a new model and the kinetics dataset.
\newblock 2017.

\bibitem[Chen et~al.(2024)Chen, Monso, Du, Simchowitz, Tedrake, and Sitzmann]{chen2024diffusion}
Boyuan Chen, Diego~Marti Monso, Yilun Du, Max Simchowitz, Russ Tedrake, and Vincent Sitzmann.
\newblock Diffusion forcing: Next-token prediction meets full-sequence diffusion.
\newblock \emph{arXiv preprint arXiv:2407.01392}, 2024.

\bibitem[Chen et~al.(2023)Chen, Xie, Chen, Hong, Li, and Yeung]{chen2023integrating}
Kai Chen, Enze Xie, Zhe Chen, Lanqing Hong, Zhenguo Li, and Dit-Yan Yeung.
\newblock Integrating geometric control into text-to-image diffusion models for high-quality detection data generation via text prompt.
\newblock \emph{arXiv preprint arXiv:2306.04607}, 2023.

\bibitem[Cheng et~al.(2023)Cheng, Liang, Shi, He, Xiao, and Li]{cheng2023layoutdiffuse}
Jiaxin Cheng, Xiao Liang, Xingjian Shi, Tong He, Tianjun Xiao, and Mu Li.
\newblock Layoutdiffuse: Adapting foundational diffusion models for layout-to-image generation.
\newblock \emph{arXiv preprint arXiv:2302.08908}, 2023.

\bibitem[Contributors(2023)]{contributors2023openscene}
O Contributors.
\newblock Openscene: The largest up-to-date 3d occupancy prediction benchmark in autonomous driving, 2023.

\bibitem[Cordts et~al.(2016)Cordts, Omran, Ramos, Rehfeld, Enzweiler, Benenson, Franke, Roth, and Schiele]{cordts2016cityscapes}
Marius Cordts, Mohamed Omran, Sebastian Ramos, Timo Rehfeld, Markus Enzweiler, Rodrigo Benenson, Uwe Franke, Stefan Roth, and Bernt Schiele.
\newblock The cityscapes dataset for semantic urban scene understanding.
\newblock In \emph{Proceedings of the IEEE conference on computer vision and pattern recognition}, pages 3213--3223, 2016.

\bibitem[Deng et~al.(2024)Deng, Tucker, Li, Guibas, Snavely, and Wetzstein]{deng2024streetscapes}
Boyang Deng, Richard Tucker, Zhengqi Li, Leonidas Guibas, Noah Snavely, and Gordon Wetzstein.
\newblock Streetscapes: Large-scale consistent street view generation using autoregressive video diffusion.
\newblock In \emph{ACM SIGGRAPH 2024 Conference Papers}, pages 1--11, 2024.

\bibitem[Dhariwal and Nichol(2021)]{dhariwal2021diffusion}
Prafulla Dhariwal and Alexander Nichol.
\newblock Diffusion models beat gans on image synthesis.
\newblock \emph{Advances in neural information processing systems}, 34:\penalty0 8780--8794, 2021.

\bibitem[Dosovitskiy et~al.(2017)Dosovitskiy, Ros, Codevilla, Lopez, and Koltun]{dosovitskiy2017carla}
Alexey Dosovitskiy, German Ros, Felipe Codevilla, Antonio Lopez, and Vladlen Koltun.
\newblock Carla: An open urban driving simulator.
\newblock In \emph{Conference on robot learning}, pages 1--16. PMLR, 2017.

\bibitem[Du et~al.(2024)Du, Yang, Dai, Dai, Nachum, Tenenbaum, Schuurmans, and Abbeel]{du2024learning}
Yilun Du, Sherry Yang, Bo Dai, Hanjun Dai, Ofir Nachum, Josh Tenenbaum, Dale Schuurmans, and Pieter Abbeel.
\newblock Learning universal policies via text-guided video generation.
\newblock \emph{Advances in Neural Information Processing Systems}, 36, 2024.

\bibitem[Esser et~al.(2021)Esser, Rombach, and Ommer]{esser2021taming}
Patrick Esser, Robin Rombach, and Bjorn Ommer.
\newblock Taming transformers for high-resolution image synthesis.
\newblock In \emph{Proceedings of the IEEE/CVF conference on computer vision and pattern recognition}, pages 12873--12883, 2021.

\bibitem[Gao et~al.(2023{\natexlab{a}})Gao, Han, Zhang, Lin, Geng, Zhou, Zhang, Lu, He, Yue, et~al.]{gao2023llamaadapterv2}
Peng Gao, Jiaming Han, Renrui Zhang, Ziyi Lin, Shijie Geng, Aojun Zhou, Wei Zhang, Pan Lu, Conghui He, Xiangyu Yue, et~al.
\newblock {LLaMA-Adapter} v2: Parameter-efficient visual instruction model.
\newblock \emph{arXiv preprint arXiv:2304.15010}, 2023{\natexlab{a}}.

\bibitem[Gao et~al.(2023{\natexlab{b}})Gao, Chen, Xie, Hong, Li, Yeung, and Xu]{gao2023magicdrive}
Ruiyuan Gao, Kai Chen, Enze Xie, Lanqing Hong, Zhenguo Li, Dit-Yan Yeung, and Qiang Xu.
\newblock Magicdrive: Street view generation with diverse 3d geometry control.
\newblock \emph{arXiv preprint arXiv:2310.02601}, 2023{\natexlab{b}}.

\bibitem[Gao et~al.(2024)Gao, Yang, Chen, Chitta, Qiu, Geiger, Zhang, and Li]{gao2024vista}
Shenyuan Gao, Jiazhi Yang, Li Chen, Kashyap Chitta, Yihang Qiu, Andreas Geiger, Jun Zhang, and Hongyang Li.
\newblock Vista: A generalizable driving world model with high fidelity and versatile controllability.
\newblock \emph{arXiv preprint arXiv:2405.17398}, 2024.

\bibitem[Geiger et~al.(2015)Geiger, Lenz, Stiller, and Urtasun]{geiger2015kitti}
Andreas Geiger, Philip Lenz, Christoph Stiller, and Raquel Urtasun.
\newblock The kitti vision benchmark suite.
\newblock \emph{URL http://www. cvlibs. net/datasets/kitti}, 2\penalty0 (5):\penalty0 1--13, 2015.

\bibitem[Guo et~al.(2023)Guo, Yang, Rao, Wang, Qiao, Lin, and Dai]{guo2023animatediff}
Yuwei Guo, Ceyuan Yang, Anyi Rao, Yaohui Wang, Yu Qiao, Dahua Lin, and Bo Dai.
\newblock Animatediff: Animate your personalized text-to-image diffusion models without specific tuning.
\newblock \emph{arXiv preprint arXiv:2307.04725}, 2023.

\bibitem[He et~al.(2022)He, Yang, Zhang, Shan, and Chen]{he2022lvdm}
Yingqing He, Tianyu Yang, Yong Zhang, Ying Shan, and Qifeng Chen.
\newblock Latent video diffusion models for high-fidelity long video generation.
\newblock 2022.

\bibitem[Heusel et~al.(2017)Heusel, Ramsauer, Unterthiner, Nessler, and Hochreiter]{Heusel2017FID}
Martin Heusel, Hubert Ramsauer, Thomas Unterthiner, Bernhard Nessler, and Sepp Hochreiter.
\newblock Gans trained by a two time-scale update rule converge to a local nash equilibrium.
\newblock 2017.

\bibitem[Ho and Salimans(2021)]{Ho2021classifierfree}
Jonathan Ho and Tim Salimans.
\newblock Classifier-free diffusion guidance.
\newblock 2021.

\bibitem[Ho et~al.(2020)Ho, Jain, and Abbeel]{ho2020denoising}
Jonathan Ho, Ajay Jain, and Pieter Abbeel.
\newblock Denoising diffusion probabilistic models.
\newblock \emph{Advances in neural information processing systems}, 33:\penalty0 6840--6851, 2020.

\bibitem[Hu et~al.(2023)Hu, Russell, Yeo, Murez, Fedoseev, Kendall, Shotton, and Corrado]{hu2023gaia}
Anthony Hu, Lloyd Russell, Hudson Yeo, Zak Murez, George Fedoseev, Alex Kendall, Jamie Shotton, and Gianluca Corrado.
\newblock Gaia-1: A generative world model for autonomous driving.
\newblock \emph{arXiv preprint arXiv:2309.17080}, 2023.

\bibitem[Jia et~al.(2023)Jia, Mao, Liu, Zhao, Wen, Zhang, Zhang, and Wang]{jia2023adriver}
Fan Jia, Weixin Mao, Yingfei Liu, Yucheng Zhao, Yuqing Wen, Chi Zhang, Xiangyu Zhang, and Tiancai Wang.
\newblock Adriver-i: A general world model for autonomous driving.
\newblock \emph{arXiv preprint arXiv:2311.13549}, 2023.

\bibitem[Kay et~al.(2017)Kay, Carreira, Simonyan, Zhang, Hillier, Vijayanarasimhan, Viola, Green, Back, Natsev, et~al.]{kay2017kinetics}
Will Kay, Joao Carreira, Karen Simonyan, Brian Zhang, Chloe Hillier, Sudheendra Vijayanarasimhan, Fabio Viola, Tim Green, Trevor Back, Paul Natsev, et~al.
\newblock The kinetics human action video dataset.
\newblock \emph{arXiv preprint arXiv:1705.06950}, 2017.

\bibitem[Kim et~al.(2019)Kim, Misu, Chen, Tawari, and Canny]{kim2019grounding}
Jinkyu Kim, Teruhisa Misu, Yi-Ting Chen, Ashish Tawari, and John Canny.
\newblock Grounding human-to-vehicle advice for self-driving vehicles.
\newblock In \emph{Proceedings of the IEEE/CVF conference on computer vision and pattern recognition}, pages 10591--10599, 2019.

\bibitem[Kim et~al.(2021)Kim, Philion, Torralba, and Fidler]{kim2021drivegan}
Seung~Wook Kim, Jonah Philion, Antonio Torralba, and Sanja Fidler.
\newblock Drivegan: Towards a controllable high-quality neural simulation.
\newblock In \emph{Proceedings of the IEEE/CVF Conference on Computer Vision and Pattern Recognition}, pages 5820--5829, 2021.

\bibitem[Kim et~al.(2024)Kim, Na, Park, Jang, Kim, Kang, and Moon]{kim2024training}
Yeongmin Kim, Byeonghu Na, Minsang Park, JoonHo Jang, Dongjun Kim, Wanmo Kang, and Il-Chul Moon.
\newblock Training unbiased diffusion models from biased dataset.
\newblock \emph{arXiv preprint arXiv:2403.01189}, 2024.

\bibitem[Li et~al.(2023{\natexlab{a}})Li, Ling, Kar, Acuna, Kim, Kreis, Torralba, and Fidler]{li2023dreamteacher}
Daiqing Li, Huan Ling, Amlan Kar, David Acuna, Seung~Wook Kim, Karsten Kreis, Antonio Torralba, and Sanja Fidler.
\newblock Dreamteacher: Pretraining image backbones with deep generative models.
\newblock In \emph{Proceedings of the IEEE/CVF International Conference on Computer Vision}, pages 16698--16708, 2023{\natexlab{a}}.

\bibitem[Li et~al.(2023{\natexlab{b}})Li, Li, Savarese, and Hoi]{li2023blip2}
Junnan Li, Dongxu Li, Silvio Savarese, and Steven Hoi.
\newblock {BLIP-2}: Bootstrapping language-image pre-training with frozen image encoders and large language models.
\newblock 2023{\natexlab{b}}.

\bibitem[Li et~al.(2022)Li, Peng, Feng, Zhang, Xue, and Zhou]{li2022metadrive}
Quanyi Li, Zhenghao Peng, Lan Feng, Qihang Zhang, Zhenghai Xue, and Bolei Zhou.
\newblock Metadrive: Composing diverse driving scenarios for generalizable reinforcement learning.
\newblock \emph{IEEE transactions on pattern analysis and machine intelligence}, 45\penalty0 (3):\penalty0 3461--3475, 2022.

\bibitem[Li et~al.(2023{\natexlab{c}})Li, Peng, Feng, Duan, Mo, Zhou, et~al.]{li2023scenarionet}
Quanyi Li, Zhenghao Peng, Lan Feng, Chenda Duan, Wenjie Mo, Bolei Zhou, et~al.
\newblock Scenarionet: Open-source platform for large-scale traffic scenario simulation and modeling.
\newblock \emph{arXiv preprint arXiv:2306.12241}, 2023{\natexlab{c}}.

\bibitem[Li et~al.(2023{\natexlab{d}})Li, Zhang, and Ye]{li2023drivingdiffusion}
Xiaofan Li, Yifu Zhang, and Xiaoqing Ye.
\newblock Drivingdiffusion: Layout-guided multi-view driving scene video generation with latent diffusion model.
\newblock \emph{arXiv preprint arXiv:2310.07771}, 2023{\natexlab{d}}.

\bibitem[Li et~al.(2023{\natexlab{e}})Li, Liu, Wu, Mu, Yang, Gao, Li, and Lee]{li2023gligen}
Yuheng Li, Haotian Liu, Qingyang Wu, Fangzhou Mu, Jianwei Yang, Jianfeng Gao, Chunyuan Li, and Yong~Jae Lee.
\newblock Gligen: Open-set grounded text-to-image generation.
\newblock In \emph{Proceedings of the IEEE/CVF Conference on Computer Vision and Pattern Recognition}, pages 22511--22521, 2023{\natexlab{e}}.

\bibitem[Liang et~al.(2022)Liang, Xie, Yu, Xia, Lin, Wang, Tang, Wang, and Tang]{liang2022bevfusion}
Tingting Liang, Hongwei Xie, Kaicheng Yu, Zhongyu Xia, Zhiwei Lin, Yongtao Wang, Tao Tang, Bing Wang, and Zhi Tang.
\newblock Bevfusion: A simple and robust lidar-camera fusion framework.
\newblock \emph{Advances in Neural Information Processing Systems}, 35:\penalty0 10421--10434, 2022.

\bibitem[Liu et~al.(2023{\natexlab{a}})Liu, Zhou, Zhu, Chang, and Guo]{liu2023apr}
Quan Liu, Yunsong Zhou, Hongzi Zhu, Shan Chang, and Minyi Guo.
\newblock Apr: online distant point cloud registration through aggregated point cloud reconstruction.
\newblock \emph{arXiv preprint arXiv:2305.02893}, 2023{\natexlab{a}}.

\bibitem[Liu et~al.(2023{\natexlab{b}})Liu, Zhu, Zhou, Li, Chang, and Guo]{liu2023density}
Quan Liu, Hongzi Zhu, Yunsong Zhou, Hongyang Li, Shan Chang, and Minyi Guo.
\newblock Density-invariant features for distant point cloud registration.
\newblock In \emph{Proceedings of the IEEE/CVF International Conference on Computer Vision}, pages 18215--18225, 2023{\natexlab{b}}.

\bibitem[Loshchilov and Hutter(2018)]{loshchilov2018fixing}
Ilya Loshchilov and Frank Hutter.
\newblock Fixing weight decay regularization in adam.
\newblock 2018.

\bibitem[Lu et~al.(2023)Lu, Huang, Zhang, Yang, and Zhang]{lu2023wovogen}
Jiachen Lu, Ze Huang, Jiahui Zhang, Zeyu Yang, and Li Zhang.
\newblock Wovogen: World volume-aware diffusion for controllable multi-camera driving scene generation.
\newblock \emph{arXiv preprint arXiv:2312.02934}, 2023.

\bibitem[Meng et~al.(2021)Meng, He, Song, Song, Wu, Zhu, and Ermon]{meng2021sdedit}
Chenlin Meng, Yutong He, Yang Song, Jiaming Song, Jiajun Wu, Jun-Yan Zhu, and Stefano Ermon.
\newblock Sdedit: Guided image synthesis and editing with stochastic differential equations.
\newblock \emph{arXiv preprint arXiv:2108.01073}, 2021.

\bibitem[Meng et~al.(2023)Meng, Rombach, Gao, Kingma, Ermon, Ho, and Salimans]{meng2023distillation}
Chenlin Meng, Robin Rombach, Ruiqi Gao, Diederik Kingma, Stefano Ermon, Jonathan Ho, and Tim Salimans.
\newblock On distillation of guided diffusion models.
\newblock In \emph{Proceedings of the IEEE/CVF Conference on Computer Vision and Pattern Recognition}, pages 14297--14306, 2023.

\bibitem[Mou et~al.(2024)Mou, Wang, Xie, Wu, Zhang, Qi, and Shan]{mou2024t2i}
Chong Mou, Xintao Wang, Liangbin Xie, Yanze Wu, Jian Zhang, Zhongang Qi, and Ying Shan.
\newblock T2i-adapter: Learning adapters to dig out more controllable ability for text-to-image diffusion models.
\newblock In \emph{Proceedings of the AAAI Conference on Artificial Intelligence}, pages 4296--4304, 2024.

\bibitem[Mullick et~al.(2023)Mullick, Jain, Gupta, and Kale]{mullick2023domain}
Koustav Mullick, Harshil Jain, Sanchit Gupta, and Amit~Arvind Kale.
\newblock Domain adaptation of synthetic driving datasets for real-world autonomous driving.
\newblock \emph{arXiv preprint arXiv:2302.04149}, 2023.

\bibitem[Nichol et~al.(2021)Nichol, Dhariwal, Ramesh, Shyam, Mishkin, McGrew, Sutskever, and Chen]{nichol2021glide}
Alex Nichol, Prafulla Dhariwal, Aditya Ramesh, Pranav Shyam, Pamela Mishkin, Bob McGrew, Ilya Sutskever, and Mark Chen.
\newblock Glide: Towards photorealistic image generation and editing with text-guided diffusion models.
\newblock \emph{arXiv preprint arXiv:2112.10741}, 2021.

\bibitem[Peng et~al.(2022)Peng, Mo, Duan, Li, and Zhou]{peng2022reward}
Zhenghao Peng, Wenjie Mo, Chenda Duan, Quanyi Li, and Bolei Zhou.
\newblock Reward-free policy learning through active human involvement.
\newblock 2022.

\bibitem[Podell et~al.(2023)Podell, English, Lacey, Blattmann, Dockhorn, M{\"u}ller, Penna, and Rombach]{podell2023sdxl}
Dustin Podell, Zion English, Kyle Lacey, Andreas Blattmann, Tim Dockhorn, Jonas M{\"u}ller, Joe Penna, and Robin Rombach.
\newblock Sdxl: Improving latent diffusion models for high-resolution image synthesis.
\newblock \emph{arXiv preprint arXiv:2307.01952}, 2023.

\bibitem[Qi et~al.(2019)Qi, Liu, Chen, and Jia]{qi20193d}
Xiaojuan Qi, Zhengzhe Liu, Qifeng Chen, and Jiaya Jia.
\newblock 3d motion decomposition for rgbd future dynamic scene synthesis.
\newblock In \emph{Proceedings of the IEEE/CVF Conference on Computer Vision and Pattern Recognition}, pages 7673--7682, 2019.

\bibitem[Qian et~al.(2023)Qian, Chen, Zhuo, Jiao, and Jiang]{qian2023nuscenes}
Tianwen Qian, Jingjing Chen, Linhai Zhuo, Yang Jiao, and Yu-Gang Jiang.
\newblock {NuScenes-QA}: A multi-modal visual question answering benchmark for autonomous driving scenario.
\newblock \emph{arXiv preprint arXiv:2305.14836}, 2023.

\bibitem[Qin et~al.(2023)Qin, Zhang, Yu, Feng, Yang, Zhou, Wang, Niebles, Xiong, Savarese, et~al.]{qin2023unicontrol}
Can Qin, Shu Zhang, Ning Yu, Yihao Feng, Xinyi Yang, Yingbo Zhou, Huan Wang, Juan~Carlos Niebles, Caiming Xiong, Silvio Savarese, et~al.
\newblock Unicontrol: A unified diffusion model for controllable visual generation in the wild.
\newblock \emph{arXiv preprint arXiv:2305.11147}, 2023.

\bibitem[Radford et~al.(2021)Radford, Kim, Hallacy, Ramesh, Goh, Agarwal, Sastry, Askell, Mishkin, Clark, et~al.]{radford2021learning}
Alec Radford, Jong~Wook Kim, Chris Hallacy, Aditya Ramesh, Gabriel Goh, Sandhini Agarwal, Girish Sastry, Amanda Askell, Pamela Mishkin, Jack Clark, et~al.
\newblock Learning transferable visual models from natural language supervision.
\newblock In \emph{International conference on machine learning}, pages 8748--8763. PMLR, 2021.

\bibitem[Raffel et~al.(2020)Raffel, Shazeer, Roberts, Lee, Narang, Matena, Zhou, Li, and Liu]{2020t5}
Colin Raffel, Noam Shazeer, Adam Roberts, Katherine Lee, Sharan Narang, Michael Matena, Yanqi Zhou, Wei Li, and Peter~J. Liu.
\newblock Exploring the limits of transfer learning with a unified text-to-text transformer.
\newblock 2020.

\bibitem[Ramesh et~al.(2022)Ramesh, Dhariwal, Nichol, Chu, and Chen]{ramesh2022hierarchical}
Aditya Ramesh, Prafulla Dhariwal, Alex Nichol, Casey Chu, and Mark Chen.
\newblock Hierarchical text-conditional image generation with clip latents.
\newblock \emph{arXiv preprint arXiv:2204.06125}, 1\penalty0 (2):\penalty0 3, 2022.

\bibitem[Richter et~al.(2016)Richter, Vineet, Roth, and Koltun]{richter2016playing}
Stephan~R Richter, Vibhav Vineet, Stefan Roth, and Vladlen Koltun.
\newblock Playing for data: Ground truth from computer games.
\newblock In \emph{Computer Vision--ECCV 2016: 14th European Conference, Amsterdam, The Netherlands, October 11-14, 2016, Proceedings, Part II 14}, pages 102--118. Springer, 2016.

\bibitem[Richter et~al.(2022)Richter, AlHaija, and Koltun]{richter2022enhancing}
Stephan~R Richter, Hassan~Abu AlHaija, and Vladlen Koltun.
\newblock Enhancing photorealism enhancement.
\newblock \emph{IEEE Transactions on Pattern Analysis and Machine Intelligence}, 45\penalty0 (2):\penalty0 1700--1715, 2022.

\bibitem[Rombach et~al.(2022)Rombach, Blattmann, Lorenz, Esser, and Ommer]{Rombach_2022_CVPR}
Robin Rombach, Andreas Blattmann, Dominik Lorenz, Patrick Esser, and Bj\"orn Ommer.
\newblock High-resolution image synthesis with latent diffusion models.
\newblock In \emph{Proceedings of the IEEE/CVF Conference on Computer Vision and Pattern Recognition (CVPR)}, pages 10684--10695, 2022.

\bibitem[Ros et~al.(2016)Ros, Sellart, Materzynska, Vazquez, and Lopez]{ros2016synthia}
German Ros, Laura Sellart, Joanna Materzynska, David Vazquez, and Antonio~M Lopez.
\newblock The synthia dataset: A large collection of synthetic images for semantic segmentation of urban scenes.
\newblock In \emph{Proceedings of the IEEE conference on computer vision and pattern recognition}, pages 3234--3243, 2016.

\bibitem[Saharia et~al.(2022)Saharia, Chan, Saxena, Li, Whang, Denton, Ghasemipour, Gontijo~Lopes, Karagol~Ayan, Salimans, et~al.]{saharia2022photorealistic}
Chitwan Saharia, William Chan, Saurabh Saxena, Lala Li, Jay Whang, Emily~L Denton, Kamyar Ghasemipour, Raphael Gontijo~Lopes, Burcu Karagol~Ayan, Tim Salimans, et~al.
\newblock Photorealistic text-to-image diffusion models with deep language understanding.
\newblock \emph{Advances in neural information processing systems}, 35:\penalty0 36479--36494, 2022.

\bibitem[Salimans and Ho(2022)]{salimans2022progressive}
Tim Salimans and Jonathan Ho.
\newblock Progressive distillation for fast sampling of diffusion models.
\newblock \emph{arXiv preprint arXiv:2202.00512}, 2022.

\bibitem[Song et~al.(2020)Song, Meng, and Ermon]{song2020DDIM}
Jiaming Song, Chenlin Meng, and Stefano Ermon.
\newblock Denoising diffusion implicit models.
\newblock \emph{arXiv preprint arXiv:2010.02502}, 2020.

\bibitem[Sun et~al.(2020)Sun, Kretzschmar, Dotiwalla, Chouard, Patnaik, Tsui, Guo, Zhou, Chai, Caine, et~al.]{sun2020scalability}
Pei Sun, Henrik Kretzschmar, Xerxes Dotiwalla, Aurelien Chouard, Vijaysai Patnaik, Paul Tsui, James Guo, Yin Zhou, Yuning Chai, Benjamin Caine, et~al.
\newblock Scalability in perception for autonomous driving: Waymo open dataset.
\newblock In \emph{Proceedings of the IEEE/CVF conference on computer vision and pattern recognition}, pages 2446--2454, 2020.

\bibitem[Sun et~al.(2022)Sun, Segu, Postels, Wang, Van~Gool, Schiele, Tombari, and Yu]{sun2022shift}
Tao Sun, Mattia Segu, Janis Postels, Yuxuan Wang, Luc Van~Gool, Bernt Schiele, Federico Tombari, and Fisher Yu.
\newblock Shift: a synthetic driving dataset for continuous multi-task domain adaptation.
\newblock In \emph{Proceedings of the IEEE/CVF Conference on Computer Vision and Pattern Recognition}, pages 21371--21382, 2022.

\bibitem[Swerdlow et~al.(2024)Swerdlow, Xu, and Zhou]{swerdlow2024street}
Alexander Swerdlow, Runsheng Xu, and Bolei Zhou.
\newblock Street-view image generation from a bird's-eye view layout.
\newblock \emph{IEEE Robotics and Automation Letters}, 2024.

\bibitem[Unterthiner et~al.(2018)Unterthiner, Van~Steenkiste, Kurach, Marinier, Michalski, and Gelly]{Unterthiner2018FVD}
Thomas Unterthiner, Sjoerd Van~Steenkiste, Karol Kurach, Raphael Marinier, Marcin Michalski, and Sylvain Gelly.
\newblock Towards accurate generative models of video: A new metric \& challenges.
\newblock \emph{arXiv preprint arXiv:1812.01717}, 2018.

\bibitem[Von~Platen et~al.(2022)Von~Platen, Patil, Lozhkov, Cuenca, Lambert, Rasul, Davaadorj, and Wolf]{von2022diffusers}
Patrick Von~Platen, Suraj Patil, Anton Lozhkov, Pedro Cuenca, Nathan Lambert, Kashif Rasul, Mishig Davaadorj, and Thomas Wolf.
\newblock Diffusers: State-of-the-art diffusion models, 2022.

\bibitem[Wang et~al.(2023{\natexlab{a}})Wang, Zhu, Huang, Chen, and Lu]{wang2023drivedreamer}
Xiaofeng Wang, Zheng Zhu, Guan Huang, Xinze Chen, and Jiwen Lu.
\newblock Drivedreamer: Towards real-world-driven world models for autonomous driving.
\newblock \emph{arXiv preprint arXiv:2309.09777}, 2023{\natexlab{a}}.

\bibitem[Wang et~al.(2023{\natexlab{b}})Wang, He, Fan, Li, Chen, and Zhang]{wang2023driving}
Yuqi Wang, Jiawei He, Lue Fan, Hongxin Li, Yuntao Chen, and Zhaoxiang Zhang.
\newblock Driving into the future: Multiview visual forecasting and planning with world model for autonomous driving.
\newblock \emph{arXiv preprint arXiv:2311.17918}, 2023{\natexlab{b}}.

\bibitem[Wen et~al.(2023)Wen, Zhao, Liu, Jia, Wang, Luo, Zhang, Wang, Sun, and Zhang]{wen2023panacea}
Yuqing Wen, Yucheng Zhao, Yingfei Liu, Fan Jia, Yanhui Wang, Chong Luo, Chi Zhang, Tiancai Wang, Xiaoyan Sun, and Xiangyu Zhang.
\newblock Panacea: Panoramic and controllable video generation for autonomous driving.
\newblock \emph{arXiv preprint arXiv:2311.16813}, 2023.

\bibitem[Weng et~al.(2023)Weng, Man, Park, Yuan, O'Toole, and Kitani]{weng2023all}
Xinshuo Weng, Yunze Man, Jinhyung Park, Ye Yuan, Matthew O'Toole, and Kris~M Kitani.
\newblock All-in-one drive: A comprehensive perception dataset with high-density long-range point clouds.
\newblock 2023.

\bibitem[Wilson et~al.(2023)Wilson, Qi, Agarwal, Lambert, Singh, Khandelwal, Pan, Kumar, Hartnett, Pontes, et~al.]{wilson2023argoverse}
Benjamin Wilson, William Qi, Tanmay Agarwal, John Lambert, Jagjeet Singh, Siddhesh Khandelwal, Bowen Pan, Ratnesh Kumar, Andrew Hartnett, Jhony~Kaesemodel Pontes, et~al.
\newblock Argoverse 2: Next generation datasets for self-driving perception and forecasting.
\newblock \emph{arXiv preprint arXiv:2301.00493}, 2023.

\bibitem[Wu et~al.(2024)Wu, Zhao, Chen, Gu, Zhao, He, Zhou, Shou, and Shen]{wu2024datasetdm}
Weijia Wu, Yuzhong Zhao, Hao Chen, Yuchao Gu, Rui Zhao, Yefei He, Hong Zhou, Mike~Zheng Shou, and Chunhua Shen.
\newblock Datasetdm: Synthesizing data with perception annotations using diffusion models.
\newblock \emph{Advances in Neural Information Processing Systems}, 36, 2024.

\bibitem[Xie et~al.(2021)Xie, Wang, Yu, Anandkumar, Alvarez, and Luo]{xie2021segformer}
Enze Xie, Wenhai Wang, Zhiding Yu, Anima Anandkumar, Jose~M Alvarez, and Ping Luo.
\newblock Segformer: Simple and efficient design for semantic segmentation with transformers.
\newblock \emph{Advances in neural information processing systems}, 34:\penalty0 12077--12090, 2021.

\bibitem[Xu et~al.(2018)Xu, Yang, Fan, Yue, Liang, Yang, and Huang]{xu2018youtube}
Ning Xu, Linjie Yang, Yuchen Fan, Dingcheng Yue, Yuchen Liang, Jianchao Yang, and Thomas Huang.
\newblock Youtube-vos: A large-scale video object segmentation benchmark.
\newblock \emph{arXiv preprint arXiv:1809.03327}, 2018.

\bibitem[Yang et~al.(2024)Yang, Gao, Qiu, Chen, Li, Dai, Chitta, Wu, Zeng, Luo, et~al.]{yang2024generalized}
Jiazhi Yang, Shenyuan Gao, Yihang Qiu, Li Chen, Tianyu Li, Bo Dai, Kashyap Chitta, Penghao Wu, Jia Zeng, Ping Luo, et~al.
\newblock Generalized predictive model for autonomous driving.
\newblock \emph{arXiv preprint arXiv:2403.09630}, 2024.

\bibitem[Yang et~al.(2023)Yang, Ma, Peng, Guo, Lin, and Yu]{yang2023bevcontrol}
Kairui Yang, Enhui Ma, Jibin Peng, Qing Guo, Di Lin, and Kaicheng Yu.
\newblock Bevcontrol: Accurately controlling street-view elements with multi-perspective consistency via bev sketch layout.
\newblock \emph{arXiv preprint arXiv:2308.01661}, 2023.

\bibitem[Zhang et~al.(2023{\natexlab{a}})Zhang, Peng, Li, and Zhou]{zhang2023cat}
Linrui Zhang, Zhenghao Peng, Quanyi Li, and Bolei Zhou.
\newblock Cat: Closed-loop adversarial training for safe end-to-end driving.
\newblock In \emph{7th Annual Conference on Robot Learning}, 2023{\natexlab{a}}.

\bibitem[Zhang et~al.(2023{\natexlab{b}})Zhang, Rao, and Agrawala]{zhang2023adding}
Lvmin Zhang, Anyi Rao, and Maneesh Agrawala.
\newblock Adding conditional control to text-to-image diffusion models, 2023{\natexlab{b}}.

\bibitem[Zhang et~al.(2022)Zhang, Peng, and Zhou]{zhang2022learning}
Qihang Zhang, Zhenghao Peng, and Bolei Zhou.
\newblock Learning to drive by watching youtube videos: Action-conditioned contrastive policy pretraining.
\newblock In \emph{European Conference on Computer Vision}, pages 111--128. Springer, 2022.

\bibitem[Zhao et~al.(2024{\natexlab{a}})Zhao, Chen, Chen, Bao, Hao, Yuan, and Wong]{zhao2024uni}
Shihao Zhao, Dongdong Chen, Yen-Chun Chen, Jianmin Bao, Shaozhe Hao, Lu Yuan, and Kwan-Yee~K Wong.
\newblock Uni-controlnet: All-in-one control to text-to-image diffusion models.
\newblock \emph{Advances in Neural Information Processing Systems}, 36, 2024{\natexlab{a}}.

\bibitem[Zhao et~al.(2024{\natexlab{b}})Zhao, Bai, Rao, Zhou, and Lu]{zhao2024unipc}
Wenliang Zhao, Lujia Bai, Yongming Rao, Jie Zhou, and Jiwen Lu.
\newblock Unipc: A unified predictor-corrector framework for fast sampling of diffusion models.
\newblock \emph{Advances in Neural Information Processing Systems}, 36, 2024{\natexlab{b}}.

\bibitem[Zhou et~al.(2021{\natexlab{a}})Zhou, He, Zhu, Wang, Li, and Jiang]{zhou2021monocular}
Yunsong Zhou, Yuan He, Hongzi Zhu, Cheng Wang, Hongyang Li, and Qinhong Jiang.
\newblock Monocular 3d object detection: An extrinsic parameter free approach.
\newblock In \emph{Proceedings of the IEEE/CVF Conference on Computer Vision and Pattern Recognition}, pages 7556--7566, 2021{\natexlab{a}}.

\bibitem[Zhou et~al.(2021{\natexlab{b}})Zhou, He, Zhu, Wang, Li, and Jiang]{zhou2021monoef}
Yunsong Zhou, Yuan He, Hongzi Zhu, Cheng Wang, Hongyang Li, and Qinhong Jiang.
\newblock Monoef: Extrinsic parameter free monocular 3d object detection.
\newblock \emph{IEEE Transactions on Pattern Analysis and Machine Intelligence}, 44\penalty0 (12):\penalty0 10114--10128, 2021{\natexlab{b}}.

\bibitem[Zhou et~al.(2021{\natexlab{c}})Zhou, Zhu, Li, Cui, Chang, and Guo]{zhou2021tempnet}
Yunsong Zhou, Hongzi Zhu, Chunqin Li, Tiankai Cui, Shan Chang, and Minyi Guo.
\newblock Tempnet: Online semantic segmentation on large-scale point cloud series.
\newblock In \emph{Proceedings of the IEEE/CVF International Conference on Computer Vision}, pages 7118--7127, 2021{\natexlab{c}}.

\bibitem[Zhou et~al.(2022)Zhou, Liu, Zhu, Li, Chang, and Guo]{zhou2022mogde}
Yunsong Zhou, Quan Liu, Hongzi Zhu, Yunzhe Li, Shan Chang, and Minyi Guo.
\newblock Mogde: Boosting mobile monocular 3d object detection with ground depth estimation.
\newblock \emph{Advances in Neural Information Processing Systems}, 35:\penalty0 2033--2045, 2022.

\bibitem[Zhou et~al.(2023)Zhou, Zhu, Liu, Chang, and Guo]{zhou2023monoatt}
Yunsong Zhou, Hongzi Zhu, Quan Liu, Shan Chang, and Minyi Guo.
\newblock Monoatt: Online monocular 3d object detection with adaptive token transformer.
\newblock In \emph{Proceedings of the IEEE/CVF Conference on Computer Vision and Pattern Recognition}, pages 17493--17503, 2023.

\bibitem[Zhou et~al.(2024)Zhou, Huang, Bu, Zeng, Li, Qiu, Zhu, Guo, Qiao, and Li]{zhou2024embodied}
Yunsong Zhou, Linyan Huang, Qingwen Bu, Jia Zeng, Tianyu Li, Hang Qiu, Hongzi Zhu, Minyi Guo, Yu Qiao, and Hongyang Li.
\newblock Embodied understanding of driving scenarios.
\newblock \emph{arXiv preprint arXiv:2403.04593}, 2024.

\bibitem[Zhu et~al.(2022)Zhu, Wu, Zhu, Jiang, Tang, Zhang, Liu, and Loy]{zhu2022celebvhq}
Hao Zhu, Wayne Wu, Wentao Zhu, Liming Jiang, Siwei Tang, Li Zhang, Ziwei Liu, and Chen~Change Loy.
\newblock {CelebV-HQ}: A large-scale video facial attributes dataset.
\newblock In \emph{ECCV}, 2022.

\end{thebibliography}
